\documentclass[runningheads]{llncs}
\usepackage{eccv}
\usepackage{eccvabbrv}
\usepackage{graphicx}
\usepackage{booktabs}
\usepackage{multirow}
\usepackage{colortbl}
\usepackage{xcolor}
\usepackage{wrapfig}
\usepackage{tabularx}
\usepackage{array}
\usepackage{anyfontsize}
\usepackage{eso-pic}
\usepackage{tcolorbox}
\tcbuselibrary{skins,breakable,listings}
\usepackage[accsupp]{axessibility}
\usepackage{orcidlink}
\usepackage{hyperref}
\usepackage{chngcntr}
\usepackage[nameinlink,noabbrev,capitalize]{cleveref}
\crefname{figure}{Fig.}{Figs.}
\Crefname{figure}{Fig.}{Figs.}
\crefname{table}{Tab.}{Tabs.}
\Crefname{table}{Tab.}{Tabs.}
\definecolor{mygray}{gray}{0.95}
\renewcommand{\paragraph}[1]{\vspace{-1.5mm}{\flushleft\textbf{#1}}}
\newcommand{\projectleader}{\textsuperscript{\dag}}
\begin{document}

\title{AnE: Pushing the Reasoning Frontier of Multimodal LLMs via Anchor Evolution}

% \titlerunning{Anchor Evolution for Multimodal Reasoning}
\titlerunning{Preprint.}

\author{
Zehao Wang\inst{1} \and
Yihan Zeng\inst{2}\projectleader\orcidlink{0009-0001-2441-5492} \and
Zidong Gong\inst{1} \and
Yuanfan Guo\inst{3} \and
Feng Zhu\inst{1} \and
Hongzhi Zhang\inst{1}\orcidlink{0000-0001-8025-346X} \and
Wei Zhang\inst{2} \and
Wangmeng Zuo\inst{1}\orcidlink{0000-0002-3330-783X} 
}

\authorrunning{Z. Wang et al.}

\institute{
Harbin Institute of Technology \and
Huawei Noah's Ark Lab \and
Independent Researcher
}

\maketitle

\AddToShipoutPictureFG*{%
  \AtPageLowerLeft{%
    \hspace{4.8cm}%
    \raisebox{3.5cm}{\footnotesize \textsuperscript{\dag}Project leader.}%
  }%
}

\begin{abstract}
Post-training via Supervised Fine-Tuning (SFT) and Reinforcement Learning (RL) is crucial for enhancing reasoning in Multimodal Large Language Models (MLLMs), yet existing paradigms often reach a performance bottleneck due to the limitations of static data. While current methods leverage self-reflection or self-evolution to push these boundaries, they still suffer from cognitive drift and hallucinated reasoning paths caused by low-quality synthetic data. 
To address these challenges, we propose \textit{\textbf{An}}chor \textit{\textbf{E}}volution (\textbf{AnE}), a new paradigm that integrates truth-anchored data curation and model evolution, achieving faithful and steady performance gains at the reasoning frontier.
Specifically, we propose Truth Anchor Expansion, which pinpoints the model \textit{failing frontier} via trajectory rollouts and leverages ground-truth databases to retrieve high-fidelity anchors for faithful data curation. Subsequently, we introduce the Scaffold-Stripping Mechanism to internalize reasoning capabilities. This mechanism first anchors reasoning paths via scaffold-augmented supervision to mitigate the learning complexity and distribution drift of direct SFT on raw data, then leverages RL to strip the scaffold template, thereby effectively transitioning the reasoning paths into intrinsic model capabilities.
Experimental results on multimodal reasoning benchmarks show that our method substantially advances the model performance frontier, improving the base model by 10.3\% across eight multimodal benchmarks and achieving state-of-the-art results. The code will be made publicly available.

\end{abstract}

\section{Introduction}
Post-training methods, particularly Supervised Fine-Tuning (SFT) and Reinforcement Learning (RL), have become the dominant strategies for developing advanced reasoning capabilities in multimodal large language models \cite{huang2025vision,li2025perception,guo2025seed1}. Recent literature highlights a distinct division of roles between the two: SFT is primarily focused on assimilating new knowledge and reasoning patterns, while RL is more about reinforcing existing behavioral patterns, rather than introducing entirely new capabilities~\cite{chu2025sft,lu2026does,gandhi2025cognitive}.
Building on these complementary strengths, modern approaches often combine both SFT and RL techniques. This integration effectively merges knowledge acquisition with behavioral alignment, thereby improving the reliability of reasoning~\cite{guo2025deepseek,zhou2025reinforced}.

\begin{figure}[t]

    \centering
    \includegraphics[width=\linewidth]{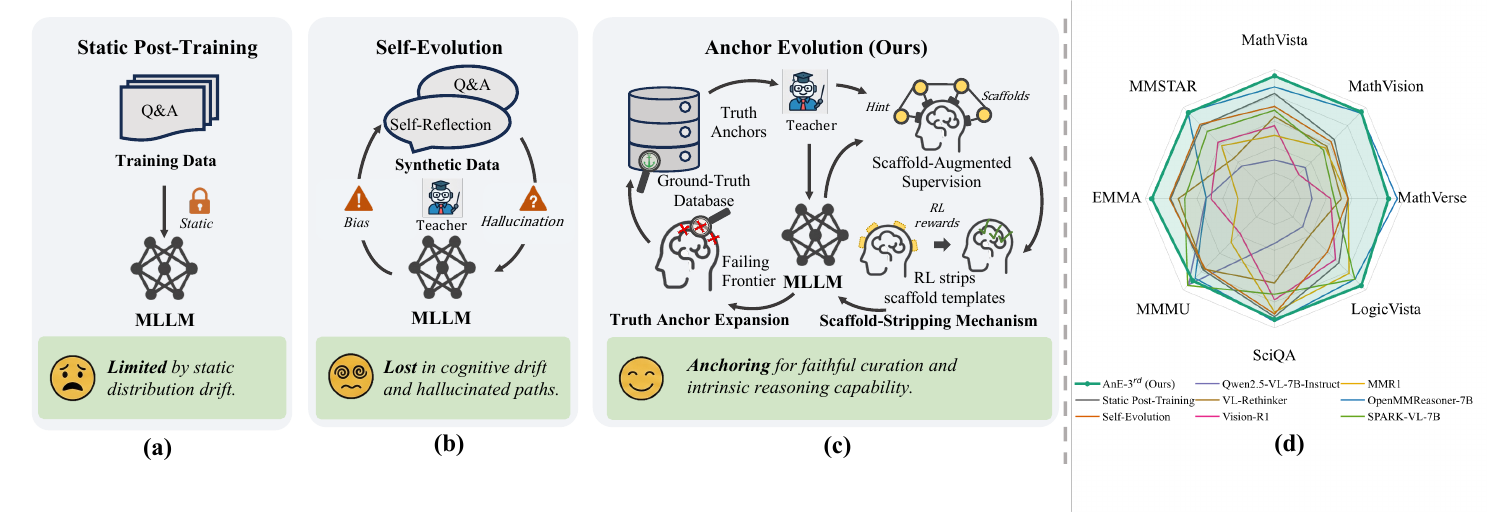} 
\caption{
\textbf{(a-c) Comparisons of training paradigms.}
Compared to static post-training and self-evolution paradigms, our proposed Anchor Evolution paradigm retrieves verified truth anchors from ground-truth databases to enable faithful data curation and internalize reasoning capabilities.
\textbf{(d) Performance comparisons.}
Anchor Evolution achieves state-of-the-art or highly competitive results across eight multimodal reasoning benchmarks.
}
    \label{fig:teaser}
\end{figure}

However, previous post-training frameworks~\cite{huang2025vision,zhang2025openmmreasoner,deng2025openvlthinker} often rely on static training data (including both SFT and RL data), as illustrated in \cref{fig:teaser}(a).
Although such data can initially improve the model’s reasoning capability to some extent, it is constrained by a fixed data distribution, which limits its ability to scale or adapt as the model’s capabilities evolve.
As a result, the model often fails to perform targeted training to address its specific weaknesses in subsequent iterations, leading to a performance bottleneck.

To break this bottleneck, recent Self-Evolution paradigms~\cite{chen2025c2,luo2025mmevol,guo2025mammoth} advocate synthetic data, as shown in \cref{fig:teaser}(b). 
For example, C2-evo \cite{chen2025c2} evolves both the model and its training data in a closed-loop manner, dynamically adjusting task complexity based on the model’s current performance. While these approaches enhance the richness and difficulty of the data, they face two major challenges. 
First, the QA pairs in the generated data often contain logical errors, which can accumulate and amplify across iterations, leading to catastrophic cognitive drift and ultimately degrading performance. 
Second, answers in synthetic data are typically sourced directly from the teacher model or human experts, which may diverge significantly from the model’s own distribution and capabilities. 
Consequently, training on such data often leads to superficial imitation of reasoning steps rather than genuine improvement in reasoning ability \cite{chen2025synergy,chen2025sft}.

Rethinking the post-training paradigm, we observe that the scalability and quality of training data play a decisive role in shaping the model’s reasoning trajectory.
While static datasets provide reliable supervision, they lack the scalability to adapt to the model’s evolving weaknesses, whereas self-evolution methods relying on synthetic data often introduce hallucinated reasoning paths and cognitive drift.
These observations suggest that effective model evolution requires both adaptability to the model’s failing frontier and reliable knowledge sources.
To this end, we introduce \textit{truth anchors}, which are verified anchor samples retrieved from ground-truth databases that serve as high-fidelity references for constructing new training data and preventing hallucinated reasoning trajectories.

Based on this observation, we seek to enable scalable, faithful model evolution. To this end, we propose \textbf{An}chor \textbf{E}volution (AnE), a post-training paradigm that grounds data evolution in verified real-world knowledge while focusing on the model's failure frontier.
The AnE framework comprises three integrated stages: 
(i) \textbf{Failing-Frontier Discovery} precisely identifies the regions where the model underperforms. This is accomplished through multiple rollouts on the seed dataset, ensuring that the training signals are closely aligned with the model’s current cognitive capabilities. A teacher model is then used to diagnose failures, extracting failure-relevant keywords and hints to guide subsequent training stages;
(ii) \textbf{Truth Anchor Expansion} retrieves verified samples from an external, real-world database to serve as high-fidelity anchors, ensuring that model training consistently relies on authentic knowledge and avoids the cognitive drift inherent in synthetic-only cycles; and
(iii) \textbf{Scaffold-Stripping Mechanism} leverages hints from an external teacher to guide the model toward correct answers and generates scaffold-augmented data for SFT. It then employs RL to remove the scaffolds, enabling the model to internalize this guidance as an intrinsic capability.
We conduct our Anchor Evolution (AnE) experiments using the Qwen2.5-VL-7B model as the base. As shown in \cref{fig:teaser}(d), across eight benchmarks, AnE demonstrates continuous improvement over multiple iterations, enhancing the base model by an average of 10.3\%. Notably, it maintains state-of-the-art performance on particularly challenging datasets such as MathVision and EMMA.

To sum up, the primary contributions of this work are summarized as follows:
\begin{itemize}

    \item We build Anchor Evolution (AnE), a post-training framework that tightly couples truth-anchored data curation and model evolution, achieving faithful and steady gains at the multimodal reasoning frontier.

    \item We propose Truth Anchor Expansion, which pinpoints where the model fails via trajectory rollouts and retrieves verified truth anchors from ground-truth databases to construct high-fidelity training data, thereby avoiding hallucinated reasoning trajectories.
    
    \item We propose scaffold-augmented supervision to anchor the model on correct reasoning paths, and then utilize RL to strip the scaffold template, enabling the model to internalize the reasoning capability.

    \item Extensive experiments on eight multimodal reasoning benchmarks show that AnE achieves consistent gains, improving the base model by 10.3\% on average and attaining state-of-the-art performance on challenging datasets such as MathVision and EMMA.

\end{itemize}

\section{Related Work}
\paragraph{Post-training for multimodal reasoning.}
Recent multimodal large language models commonly adopt a post-training pipeline that combines supervised fine-tuning (SFT) and reinforcement learning (RL) \cite{Qwen2.5-VL,guo2025deepseek,zhang2025openmmreasoner,leng2025mmr1,wei2025open,huang2025vision,liu2026automated}.
SFT is typically used to provide a cold start with structured reasoning traces, while RL methods (e.g., GRPO \cite{guo2025deepseek}) further improve robustness by optimizing task rewards.
However, in many pipelines, the SFT/RL datasets are prepared in advance, which keeps the training distribution largely static throughout optimization.
As the model evolves, such static data become increasingly misaligned with the model’s capability frontier, often resulting in diminishing returns.

\paragraph{Iterative SFT--RL optimization.}
Beyond one-pass post-training, recent work explores iterative optimization by alternating SFT and RL stages \cite{guo2025seed1,li2025xiaomi,chen2025step,ma2025learning,qiu2025metis}.
For example, OpenVLThinker \cite{deng2025openvlthinker} performs multiple rounds of SFT and RL using a manually constructed static dataset and reject sampling, gradually introducing harder training samples in later stages. Such schedules can improve stability, for instance by reducing policy drift, and help the model focus on hard samples exposed during RL. Nevertheless, the data construction in these pipelines is often static or only weakly adaptive \cite{chen2025step}, which limits sustained progress at the reasoning frontier.
In contrast, we evolve the training data with Truth Anchor Expansion to target the model's failing frontier while iterating SFT and RL.

\paragraph{Feedback-driven optimization and self-evolution.}
Feedback-driven methods introduce external critics \cite{zhang2025adhint,liu2026automated} or self-reflection \cite{li2025reflectevo} to refine reasoning processes, and self-evolution systems further form closed-loop training cycles using synthetic supervision \cite{thawakar2025evolmm,chen2025c2,koh2025adastar}.
While these approaches can improve performance, they may suffer from cognitive drift and hallucinated reasoning paths when the training signal is dominated by low-quality self-synthetic data.
Different from purely synthetic self-evolution, our Anchor Evolution uses Truth Anchor Expansion to retrieve verified truth anchors from ground-truth databases for faithful data curation, and employs the Scaffold-Stripping Mechanism (scaffold-augmented supervision followed by RL to strip the scaffold template) to internalize reasoning capability.

\section{Method}
\label{sec:method}

\begin{figure*}[t]
    \centering
    \includegraphics[width=\linewidth]{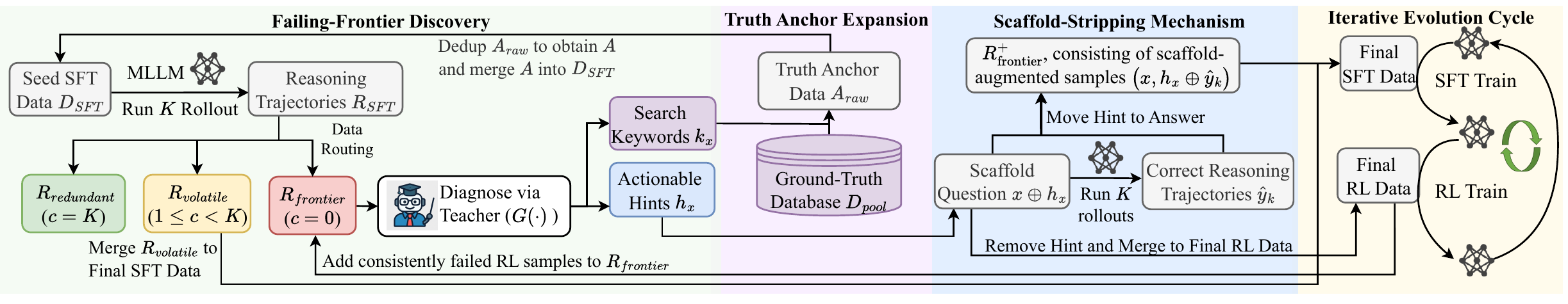} 
    \caption{\textbf{Overview of Anchor Evolution (AnE).} AnE consists of three stages: \textbf{Failing-Frontier Discovery}, where we run rollouts and use a teacher for diagnosis; \textbf{Truth Anchor Expansion}, where we retrieve verified truth anchors from ground-truth databases; and the \textbf{Scaffold-Stripping Mechanism}, where we perform scaffold-augmented SFT and then use RL to strip the scaffold template. These stages are repeated in an \textbf{Iterative Evolution Cycle} to push the model’s reasoning frontier.}
    \label{fig:pipeline}
\end{figure*}

In this section, we introduce Anchor Evolution (AnE), an iterative post-training framework designed to achieve faithful and sustained performance gains at the reasoning frontier.
As illustrated in \cref{fig:pipeline}, the AnE process comprises three stages:
(i) Failing-Frontier Discovery (\cref{subsec:3.1}), which performs multiple rollouts of the model under training to probe the capability landscape. A teacher model then provides diagnostic analysis, transforming the reasoning trajectories of hard failure samples into search keywords and actionable hints;
(ii) Truth Anchor Expansion (\cref{subsec:3.2}), which retrieves high-fidelity anchor samples from a ground-truth database using the generated search keywords for faithful data curation;
(iii) Scaffold-Stripping Mechanism (\cref{subsec:3.3}), which leverages the actionable hints to construct scaffold-augmented data for SFT, and subsequently applies RL to remove the scaffold templates, thereby transitioning the reasoning trajectory into intrinsic model capabilities. 
These three stages form an Iterative Evolution Cycle that is repeated to progressively evolve the model (\cref{subsec:3.4}).

\subsection{Failing-Frontier Discovery}
\label{subsec:3.1}

\paragraph{Preliminary RL for Initial Reasoning Ability.}
To initiate the Anchor Evolution cycle, a preliminary RL phase is conducted to establish the model’s initial reasoning state, denoted as $\mathcal{M}_{\text{PreRL}}$, using a seed RL dataset $\mathcal{D}_{\text{RL}}$.
This stage serves as a warm-up rather than the core contribution of AnE.
By incentivizing the activation of existing knowledge through preliminary RL, the model’s latent reasoning capacity is more fully exploited~\cite{meng2025mm,qiu2025metis}. 
Consequently, the model is pushed toward its intrinsic reasoning ceiling, thereby reducing the risk that subsequent failure discovery is confounded by superficial instruction-following weaknesses or elicitation artifacts.

\paragraph{Data Routing.} 
To assess the model's current performance, we perform $K$ stochastic rollouts for each sample $x$ in the seed SFT dataset $\mathcal{D}_{\text{SFT}}$, using $\mathcal{M}_{\text{PreRL}}$ as a diagnostic probe. 
This procedure produces a set of reasoning-trajectory groups
$\mathcal{R}_{\mathrm{SFT}} = \{ r_x \mid x \in \mathcal{D}_{\mathrm{SFT}} \}$,
where each group $r_x = \{\hat{y}_{x,1}, \ldots, \hat{y}_{x,K}\}$ contains the $K$ stochastic rollout trajectories generated for sample $x$.
Each prediction $\hat{y}_k$ is evaluated against the ground-truth label $y$ using an indicator function $\mathbb{I}(\hat{y}, y) \in \{0, 1\}$, which returns $1$ if the prediction is correct and $0$ otherwise. 
The model's performance on sample $x$ is then quantified by the success count:
\[
    c(x) = \sum_{k=1}^{K} \mathbb{I}(\hat{y}_k, y).
\]
Based on the success count $c(x)$, we partition $R_{\text{SFT}}$ into three mutually exclusive subsets:
(i) Redundant set ($R_{\text{redundant}}$). Samples with $c = K$, representing fully redundant signals, are categorized as $R_{\text{redundant}}$ and pruned from the evolution cycle to eliminate unnecessary computational overhead.
(ii) Volatile set ($R_{\text{volatile}}$). Samples with $1 \leq c < K$, which reflect emerging or unstable capabilities, are grouped into $R_{\text{volatile}}$. These samples are retained in the training set to preserve optimization stability and prevent premature forgetting.
(iii) Failing frontier ($R_{\text{frontier}}$): Samples with $c = 0$, representing a complete failure in the seed SFT data, constitute the failing frontier. This frontier plays a key role in Anchor Evolution, as these samples highlight critical gaps in the model's capabilities. 
In addition, we include in \(R_{\text{frontier}}\) the samples from the preliminary RL phase that the model continues to fail consistently.
These samples, like those with \( c = 0 \) in the seed SFT data, fall outside the model's exploration capabilities and cannot be resolved through standard RL methods. This strategy ensures that the failing frontier captures a more comprehensive view of the model’s most persistent and fundamental logical deficiencies.

\paragraph{Diagnostic Analysis via Teacher Model.} 
For each $r_x \in R_{\text{frontier}}$, we employ a diagnostic function $G(\cdot)$, implemented via a teacher model, to conduct a fine-grained analysis of the corresponding trajectory set ${r}_x$.
Specifically, $G(\cdot)$ first identifies the root cause of failure by categorizing errors into four distinct dimensions: knowledge, understanding, reasoning, and execution. 
Guided by this taxonomy, it then derives a metadata tuple as follows:
$$(k_x, h_x) \leftarrow G(x, y_x, \{\hat{y}_1,\ldots,\hat{y}_K\}).$$
The search keywords ${k}_x$ consist of 3-5 nouns related to $x$ and $r_x$, which act as semantic probes for retrieving truth anchors (\cref{subsec:3.2}) from an external ground-truth database, specifically associated with failure-neighborhood samples. 
Meanwhile, the actionable hints ${h}_x$ are 1-3 sentences used to help solve problem $x$, serving as structured scaffolds that reduce problem difficulty, thereby making the target solution more accessible to the model (\cref{subsec:3.3}).

\subsection{Truth Anchor Expansion}
\label{subsec:3.2}

Once the failing frontier $R_{\text{frontier}}$ is identified, a key challenge is how to expand these failure cases effectively. 
Relying solely on synthetic question generation~\cite{guo2025mammoth,luo2025mmevol} may introduce unanchored cognitive drift, in which the model’s reasoning gradually diverges from objective logical truth.
To address this issue, we introduce Truth Anchor Expansion, a retrieval-augmented mechanism that grounds the expansion process in an authentic external database. 
Specifically, AnE retrieves verifiable ground-truth samples from the database that match the model’s identified logical deficiencies. 
This helps the model learn from failure cases while maintaining factual and logical consistency.

\paragraph{Truth Anchor Retrieval for Failure Frontier.} For each failure reasoning trajectory $r_x \in R_{\text{frontier}}$, the associated search keywords $k_x$ serve as semantic probes for targeted retrieval.
We define a large-scale ground-truth database, $\mathcal{D}_{\text{pool}}$, which contains verified samples spanning diverse domains.
Based on the identified failure, the anchoring process selects 
a set of \textit{truth anchors} ${a}_x \subset \mathcal{D}_{\text{pool}}$ 
that lie in its semantic and logical neighborhood.
A pre-trained embedding model $\phi(\cdot)$ projects both 
the retrieval query $q_x = \text{Prompt}(k_x)$ 
and candidate samples into a shared latent representation space. 
Truth-anchor retrieval then reduces to a similarity-based ranking problem:
\[
{a}_{x} =
\left\{
x' \in \mathcal{D}_{\text{pool}}
\;\middle|\;
\operatorname{rank}\!\left(
\operatorname{sim}\big(\phi(q_x), \phi(x')\big)
\right) \le N
\right\},
\]
where $\operatorname{sim}(\cdot, \cdot)$ denotes cosine similarity 
and $N$ specifies the top-$N$ retrieval threshold.
The resulting anchor data is denoted as 
$\mathcal{A}_{\text{raw}} = \{ {a}_x \}$. 
This procedure ensures that retrieved samples directly address 
the identified deficiencies while preserving strict alignment 
with verified ground-truth knowledge.

\paragraph{Merging Anchor Data with the Seed SFT Dataset.} To ensure that the augmented anchor data provides sufficient learning signals, the raw anchor set, $\mathcal{A}_{\text{raw}}$, is first deduplicated to form $\mathcal{A}$. 
Subsequently, $\mathcal{A}$ is integrated into the fine-tuning dataset, $\mathcal{D}_{\text{SFT}}$, and undergoes $K$ stochastic rollouts and data partitioning as described in \cref{subsec:3.1}.
Redundant data is discarded, and only the volatile data, denoted as $R_{\text{volatile}}$, is retained in the training set. The resulting frontier, $R_{\text{frontier}}$, is then directed into the process outlined in \cref{subsec:3.3}.

\subsection{Scaffold-Stripping Mechanism}
\label{subsec:3.3}
After Failing-Frontier Discovery and Truth Anchor Expansion, we obtain \( R_{\text{volatile}} \) and \( R_{\text{frontier}} \). 
\( R_{\text{volatile}} \) consists of partial samples that represent reasoning trajectories successfully generated by the model, which can be directly utilized for SFT training.
In contrast, \( R_{\text{frontier}} \) still lacks correct reasoning trajectories. 
While current methods, such as teacher distillation or the use of reasoning trajectory data annotated by human experts, can be employed, these datasets often differ significantly from the model's current capabilities.
As a result, directly applying them for SFT tends to lead to superficial imitation rather than enabling the model to truly master and generate high-quality reasoning abilities.
To address this, we leverage the actionable hint from~\cref{subsec:3.1} to construct Scaffold-Augmented Data for \( R_{\text{frontier}} \), which is then used for SFT training. 
Subsequently, the hints from \( R_{\text{frontier}} \) are removed and integrated into the RL dataset to effectively transform the assisted reasoning paths into intrinsic model capabilities.

\paragraph{Scaffold-Augmented SFT.} 
To bridge the gap between the failure frontier and model ability, we leverage actionable hints \( h_x \) to elicit successful reasoning trajectories from the model's own distribution. 
Specifically, for each \( x \), we first construct a guided prompt by appending the hint to the question: 
\[
x_{\text{scaffold}} = x \oplus h_x.
\]
We then execute \( K \) stochastic rollouts.
For every successful trajectory \( \hat{y}_k \) that reaches the correct ground truth (i.e., where \( 1 \le c \le K \) under guidance), we relocate the scaffold \( h_x \) from the input to the prefix of the output. This transformation yields the following training tuple:
\[
(x, h_x \oplus \hat{y}_k).
\]
This process transforms \( R_{\text{frontier}} \) into  \( R_{\text{frontier}}^{+} \) that contains the correct reasoning trajectories. Finally, we extract the correct trajectories from both \( R_{\text{frontier}}^{+} \) and \( R_{\text{volatile}} \), using them for Final SFT training.

AnE utilizes reasoning paths that the model either discovers independently or is guided to discover through teacher hints. This ensures that the model consistently focuses on the boundary of its capabilities, learning from data where its abilities are unstable, thus stabilizing the training process. It provides a solid foundation for the model to eventually master these tasks independently.

\paragraph{Internalization via RL Stripping.} 
While SFT enables the model to solve previously impossible problems through actionable hints from a teacher, the model remains externally dependent. To achieve true capability expansion, we utilize RL as a mechanism for scaffold stripping.
During this phase, we add \( R_{\text{frontier}}^{+} \) to the RL dataset (without the actionable hint in the input question). 
To ensure the RL dataset contains sufficient learning signals, similar to the seed SFT data in \cref{subsec:3.1}, we also perform \( K \) rollouts on it and filter out the samples where \( c = K \), forming the final RL dataset.
Finally, we perform RL training on the final RL dataset. The model is presented only with the original problem \( x \) and must independently reach the correct answer \( y \) without any hints. 

This optimization bridges the gap between hint-aided success and independent mastery, compelling the model to recover the underlying reasoning logic autonomously.
As a result, the previously scaffold-dependent capabilities become solidified in the model's parametric weights, successfully advancing the frontier of the model's reasoning ability.

\subsection{Iterative Evolution Cycle}
\label{subsec:3.4}
The subsequent process of the AnE framework follows a continuous SFT-RL evolution cycle. 
At the beginning of each new iteration, the framework takes the model from the previous round as its starting point and re-executes the procedures outlined in \cref{subsec:3.1} to \cref{subsec:3.3}, excluding the preliminary RL phase. 
This cycle is repeated to continuously advance the model at the reasoning frontier.
\section{Experiments}
\subsection{Experimental Settings}

\textbf{Implementation Details.} 
The SFT process of the AnE framework is implemented using MindSpeed-MM, while the RL phase is conducted with GSPO~\cite{zheng2025group} on veRL~\cite{sheng2024hybridflow}. For data routing, $K = 4$ stochastic rollouts are performed. The teacher model, Qwen2.5-VL-72B-Instruct~\cite{Qwen2.5-VL}, is used to conduct diagnostic analysis, while Qwen3-Embedding-8B~\cite{zhang2025qwen3} is employed for embedding search keywords and data in the external database during truth anchor expansion. For truth anchor retrieval, $N = 5$ nearest neighbors are selected.
The model curriculum begins with a preliminary RL phase to establish the initial reasoning ability. It then undergoes three complete evolution iterations, each consisting of an SFT stage with scaffolded assistance, followed by an RL stage for scaffold shedding. Training continues until either the SFT loss or the RL mean reward converges.
All experiments are performed on clusters of 64 Ascend 910B NPUs, with Qwen2.5-VL-7B-Instruct as the base model. Detailed hyperparameters are provided in the Appendix.

\paragraph{Training datasets.} 
The evolution cycle starts with a seed RL dataset of 80K samples and a seed SFT dataset of 500K samples. 
The seed SFT dataset is derived from a diverse set of multimodal reasoning and mathematical benchmarks, including LLaVA-CoT~\cite{xu2025llava}, OpenVLThinker~\cite{deng2025openvlthinker}, We-Math2.0~\cite{qiao2025we2}, Mulberry~\cite{yao2024mulberry}, MMR1-SFT~\cite{leng2025mmr1}, and MiroMind-M1~\cite{li2025miromind}. 
For the seed RL dataset, a reasoning-focused benchmark suite is curated, emphasizing answer verifiability and structured outputs, including AI2D~\cite{kembhavi2016diagram}, Geometry3k~\cite{lu2021inter}, MMEureka~\cite{meng2025mm}, ViRL~\cite{wang2025vl}, TQA~\cite{kembhavi2017you}, We-Math~\cite{qiao2025we}, PuzzleVQA~\cite{chia2024puzzlevqa}, AlgoPuzzleVQA~\cite{ghosal2024language}, ThinkLiteVL~\cite{wang2025sota}, and MMR1-RL~\cite{leng2025mmr1}. 
The external database \( D_{pool} \) is constructed using FineVision~\cite{wiedmann2025finevision}, a multimodal dataset comprising 24 million samples. An image deduplication pipeline is employed for FineVision across 66 multimodal datasets from lmms-eval~\cite{lmmseval} to prevent data leakage.

\paragraph{Baseline and Benchmarks.} 
Performance is evaluated across a suite of multimodal reasoning benchmarks, including MathVista(testmini)~\cite{lu2023mathvista}, MathVision~\cite{wang2024measuring}, MathVerse(testmini)~\cite{zhang2024mathverse}, LogicVista~\cite{xiao2024logicvista}, ScienceQA~\cite{saikh2022scienceqa}, MMMU(val)~\cite{yue2024mmmu}, EMMA(all)~\cite{hao2025can}, and MMSTAR~\cite{chen2024we}. Top-1 Accuracy (\%) is reported using greedy decoding to ensure reproducibility.

\subsection{Main Results}
\cref{tab:main-results} presents a comprehensive comparison between AnE and representative baselines across a wide range of multimodal reasoning benchmarks.

\paragraph{Comparison with Base Model.}
AnE-$1^{st}$ achieves a significant improvement of 8.3\% on Qwen2.5-VL-7B-Instruct, demonstrating that AnE effectively identifies the Failure Frontier in the seed dataset, expands it with targeted samples from the ground-truth dataset, and simplifies problem difficulty through the Scaffold-Stripping mechanism, all while successfully internalizing guidance from the Teacher Model.

\paragraph{Continuous Improvement.} 
AnE-$2^{nd}$ and AnE-$3^{rd}$ show further improvements of 0.6 and 1.4, respectively, compared to the previous rounds. Especially on the more challenging benchmark tasks, MathVision and EMMA, AnE-$2^{nd}$ and AnE-$3^{rd}$ continue to deliver incremental gains. This highlights the method's ability to consistently identify weaknesses in the model’s capabilities and drive progress at the forefront of reasoning.

\paragraph{Comparison with Related Methods.} To ensure a fair comparison, we primarily benchmark against recent multimodal reasoning methods built upon the same 7B-parameter base model.
First, compared to static data methods, such as one-pass SFT+RL approaches (e.g., OpenMMReasoner\cite{zhang2025openmmreasoner} and Vision-R1\cite{huang2025vision}) and multi-round methods with static data (e.g., OpenVLThinker), our method demonstrates superior performance, highlighting that dynamic data expansion is a more effective strategy for enhancing model capabilities. 
In addition, AnE outperforms synthetic data methods like C2-Evo\cite{chen2025c2}, further proving its greater effectiveness. 
Finally, it surpasses other reflective approaches, including SRPO\cite{wan2025srpo} and ADHint\cite{zhang2025adhint}, showcasing that our Scaffold-Augmented Data and Internalization through Scaffold Stripping are highly effective in helping the model explore its limits and internalize guidance from external teachers.
Overall, AnE-$3^{rd}$ achieves state-of-the-art average performance across eight benchmarks.

\begin{table*}[t]
  \centering
  \small
  \caption{
\textbf{Comparison with related models on multimodal reasoning benchmarks.}
Accuracy (\%) is reported.
``--'' indicates that the result is neither reported in the original paper nor reproducible due to unavailable checkpoints.
$\dagger$ denotes results reproduced with our evaluation pipeline.
Preliminary RL serves as the warm-up phase of AnE.
The best results are shown in \textbf{bold}, and the second-best results are \underline{underlined}.
  }
  \label{tab:main-results}
  \renewcommand{\arraystretch}{1.15}
  \definecolor{mygray}{gray}{0.95}

  \resizebox{\textwidth}{!}{%
  \begin{tabular}{lcccccccc|c}
    \toprule
    \textbf{Method} & \textbf{MathVista} & \textbf{MathVision} & \textbf{MathVerse} & \textbf{LogicVista} & \textbf{SciQA} & \textbf{MMMU} & \textbf{EMMA} & \textbf{MMSTAR} & \textbf{Avg} \\
    \midrule
    Qwen2.5-VL-7B-Instruct~\cite{Qwen2.5-VL} & 68.2 & 25.1 & 49.2 & 39.3$^\dagger$ & 85.4 & \underline{58.6} & 24.6 & 62.5 & 51.6$^\dagger$ \\

    Vision-R1~\cite{huang2025vision} & 73.5$^\dagger$ & 22.8 & 52.4$^\dagger$ & 46.0$^\dagger$ & 91.4$^\dagger$ & 52.3$^\dagger$ & 23.6$^\dagger$ & 65.8$^\dagger$ & 53.5$^\dagger$ \\
    MMR1~\cite{leng2025mmr1} & 72.0 & 31.8 & 55.4 & 48.8 & 92.7$^\dagger$ & 53.4$^\dagger$ & 18.6$^\dagger$ & 65.3$^\dagger$ & 54.8$^\dagger$ \\
    C2-Evo~\cite{chen2025c2} & 68.7 & -- & -- & -- & -- & -- & -- & -- & -- \\
    Revisual-R1~\cite{chen2025advancing} & 73.1 & \textbf{48.8} & 53.6 & \textbf{52.3} & 91.9$^\dagger$ & 57.2$^\dagger$ & 19.5$^\dagger$ & 57.1$^\dagger$ & 56.7$^\dagger$  \\
    Metis-RISE~\cite{qiu2025metis} & 75.8 & 28.7 & 51.0 & 49.7 & 93.4$^\dagger$ & 55.8$^\dagger$ & 28.5$^\dagger$ & 66.6$^\dagger$ & 56.2$^\dagger$ \\
    SRPO~\cite{wan2025srpo} & 75.8 & 32.9 & 55.8 & -- & -- & 57.1 & 29.6 & -- & -- \\
    VL-Rethinker~\cite{wang2025vl} & 74.9 & 32.3 & 54.2 & 40.9$^\dagger$ & 89.6 & 56.7 & 29.7 & 63.6$^\dagger$ & 55.2$^\dagger$ \\
    SPARK-VL-7B~\cite{liu2025spark} & 75.9 & 31.1 & 53.0 & 50.0 & 90.8 & \textbf{58.7} & 28.5$^\dagger$ & 67.3 & 56.9$^\dagger$ \\
    LLaVA-Critic-R1~\cite{wang2025llava} & 74.0 & 30.6 & 49.7 & 39.5$^\dagger$ & 84.2$^\dagger$ & 55.2 & 28.3 & 65.1 & 53.3$^\dagger$ \\
    OpenVLThinker~\cite{deng2025openvlthinker} & 72.3 & 25.9 & 47.9 & 43.5$^\dagger$ & 88.9$^\dagger$ & 53.3$^\dagger$ & 26.8 & 61.6$^\dagger$ & 52.5$^\dagger$ \\

    OpenMMReasoner-7B~\cite{zhang2025openmmreasoner} & 79.5 & 43.6 & \textbf{63.8} & 50.0 & \textbf{93.7}$^\dagger$ & 57.8 & 24.5$^\dagger$ & \textbf{70.0}$^\dagger$ & 60.4$^\dagger$ \\
    ADHint~\cite{zhang2025adhint} & 74.4 & -- & 60.6 & 48.7 & -- & 56.4 & -- & -- & -- \\

    \midrule
    Preliminary RL & 77.9 & 32.8 & 53.6 & 45.5 & 92.6 & \textbf{57.7} & 30.1 & 67.2 & 57.2 \\
    \rowcolor{mygray}
    \textbf{AnE-$1^{st}$} & 79.6 & 39.7 & 60.5 & 49.1 & 93.5 & 57.7 & 30.5 & 68.3 & 59.9 \\
    \rowcolor{mygray}
    \textbf{AnE-$2^{nd}$} & \underline{80.1} & 40.9 & 60.9 & 49.8 & \underline{93.6} & 57.0 & \underline{33.0} & 68.9 & \underline{60.5} \\
    \rowcolor{mygray}
    \textbf{AnE-$3^{rd}$} & \textbf{81.2} & \underline{43.9} & \underline{62.3} & \underline{51.3} & 93.5 & 58.1 & \textbf{34.6} & \underline{69.9} & \textbf{61.9} \\
    \bottomrule
  \end{tabular}%
  }
\end{table*}

\subsection{Comparison with Prior Post-Training Paradigms}

\begin{figure}[t]
    \centering
    \begin{minipage}[t]{0.48\linewidth}
        \centering
        \includegraphics[width=\linewidth]{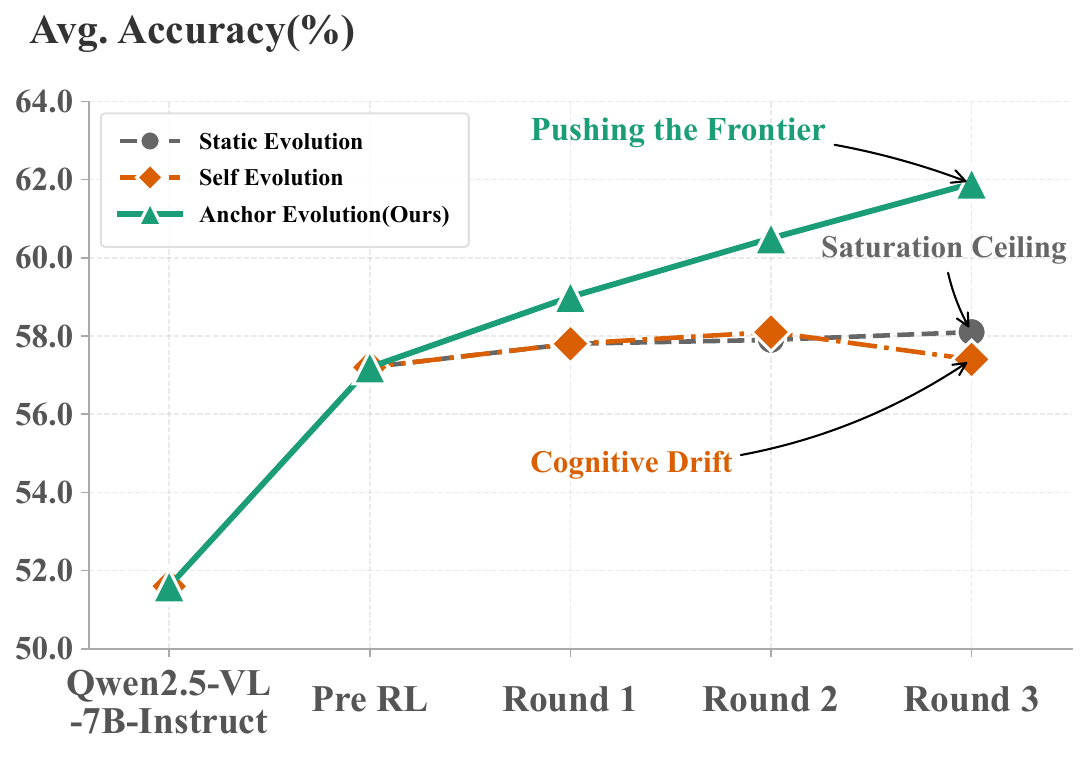}
        \vspace{-1mm}
        \centerline{\small (a) Evolution strategies}
    \end{minipage}
    \hfill
    \begin{minipage}[t]{0.48\linewidth}
        \centering
        \includegraphics[width=\linewidth]{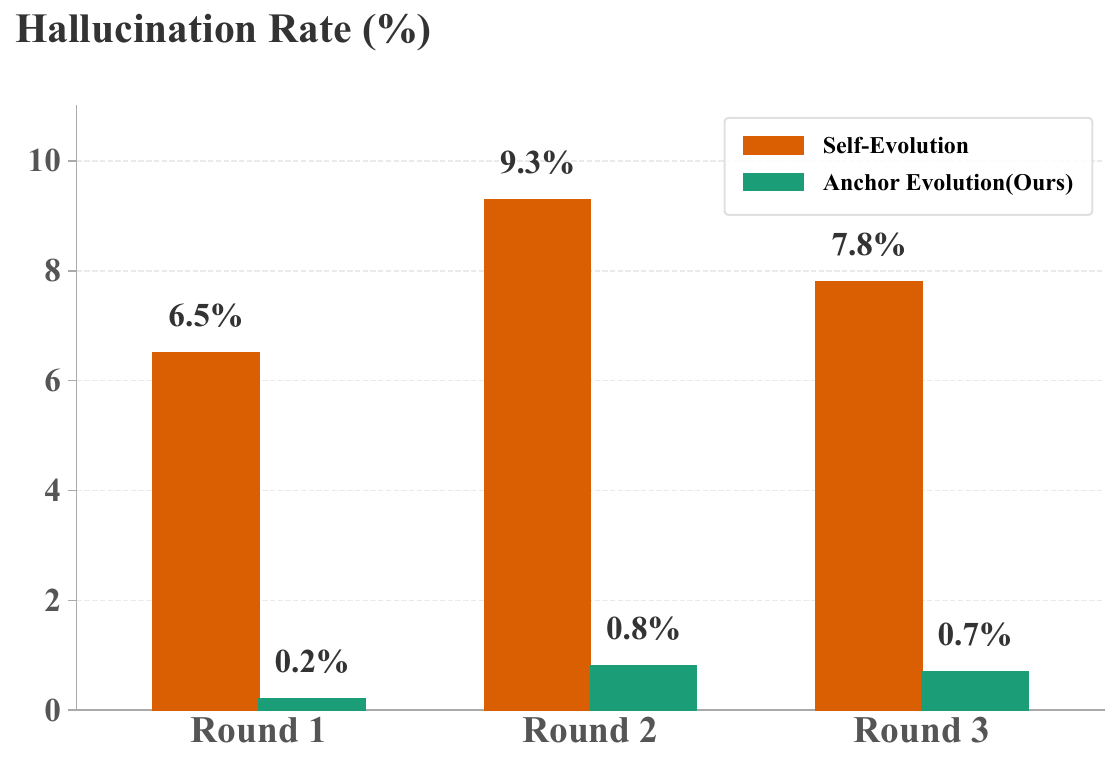}
        \vspace{-1mm}
        \centerline{\small (b) Hallucination rate}
    \end{minipage}
\caption{
\textbf{Comparison with prior evolution paradigms.}
\textbf{(a)} Static Evolution quickly saturates, Self-Evolution suffers performance degradation in later rounds, while Anchor Evolution consistently improves across rounds.
\textbf{(b)} Hallucination rates of expanded training data across different evolution rounds.
Anchor Evolution produces substantially fewer hallucinated samples than Self-Evolution.
}
    \label{fig:compare-hallucination}
\end{figure}

In this section, we conduct a fair comparison between Anchor Evolution and two commonly used iterative post-training strategies: Static Evolution and Self-Evolution.
For Static Evolution, we use the same seed SFT and RL datasets as in AnE and perform alternating SFT and RL training for three rounds, without any data augmentation or modification.
For Self-Evolution, we use Qwen2.5-VL-72B-Instruct as the teacher model to synthesize new questions and reasoning paths from the same seed SFT dataset as AnE for constructing SFT training data.
The prompts follow the settings in~\cite{guo2025mammoth,luo2025mmevol}.
For fair comparison, RL training in both Static Evolution and Self-Evolution consistently uses the same static seed RL dataset.

As shown in \cref{fig:compare-hallucination}(a), Static Evolution quickly reaches a performance plateau across multiple rounds.
In contrast, Self-Evolution achieves larger improvements than Static Evolution in Round~1 and Round~2, suggesting that synthetic data expansion can enlarge the model's capability boundary to some extent compared with purely static training.
However, a noticeable performance drop appears in Round~3, indicating that self-evolution may introduce hallucinated or low-quality synthetic data during iterative generation, which in turn affects subsequent training.
To verify this hypothesis, we sample 1K expanded samples from each round and use GPT-4o to identify hallucinated reasoning or incorrect ground-truth answers.
As shown in \cref{fig:compare-hallucination}(b), Self-Evolution has consistently higher hallucination rates than Anchor Evolution across three rounds (6.5/9.3/7.8\% vs. 0.2/0.8/0.7\%).
This confirms that synthetic expansion is more prone to noisy supervision and cognitive drift, whereas AnE's truth-anchored filtering yields more faithful data for stable evolution.

Overall, Anchor Evolution consistently improves performance across rounds and achieves better results than both Static Evolution and Self-Evolution, demonstrating its ability to provide stable and reliable capability expansion during iterative post-training.

\subsection{Ablation Study}
In this section, we perform ablation studies to assess the key design choices in AnE, including the synergy between Truth Anchor Expansion and Scaffolded Reasoning Supervision, the effectiveness and scaling behavior of Truth Anchor Expansion, and the impact of the Scaffold-Stripping Mechanism.
To isolate the effects of the proposed AnE components, we use Preliminary RL as the baseline in all ablation studies.
Preliminary RL only provides a warm-up initialization and is shared by all variants; therefore, the reported gains reflect the contributions of Truth Anchor Expansion, Scaffold-Stripping, and their iterative coupling.

\paragraph{Component of Truth Anchor Expansion and Scaffolding.} 
As shown in \cref{tab:ablation-hint-rag}, starting from the Preliminary RL baseline (57.2\% avg), incorporating Truth Anchor Expansion alone results in a significant improvement of +1.4\%, reaching 58.6\%.
This demonstrates that our Truth Anchoring method effectively expands true data at the Failure Frontier without introducing hallucinations, thereby boosting the model’s capabilities.
Similarly, adding only Anchored Scaffolding improves the baseline to 57.9\% (+0.7\% average), confirming that temporary cognitive guidance effectively bridges the capability gap, allowing external knowledge from the teacher model to be internalized into the model's own reasoning abilities.
Most importantly, the full AnE-$1^{st}$ (59.9\%) surpasses the individual gains of either component, achieving a total uplift of +2.7\% over the baseline.
This synergistic effect highlights that Truth Anchor Expansion and Scaffolding can work together effectively, pushing the boundaries of the model’s reasoning capabilities.

\begin{table*}[t]
  \centering
  \small
\caption{\textbf{Component ablation analysis.} Results show that both Truth Anchor Expansion and the Scaffold-Stripping Mechanism are effective on their own, and combining them in AnE yields the best overall performance.}
  \label{tab:ablation-hint-rag}
  \renewcommand{\arraystretch}{1.2}
  \resizebox{\textwidth}{!}{%
  \begin{tabular}{l|cccccccc|c}
    \toprule
    \textbf{Setting} & \textbf{MathVista} & \textbf{MathVision} & \textbf{MathVerse} & \textbf{LogicVista} & \textbf{SciQA} & \textbf{MMMU} & \textbf{EMMA} & \textbf{MMSTAR} & \textbf{Avg} \\
    \midrule

    Preliminary RL & 77.9 & 32.8 & 53.6 & 45.5 & 92.6 & \textbf{57.7} & 30.1 & 67.2 & 57.2 \\
     $\hookrightarrow$ w/ Truth Anchor Expansion      & 78.3 & 36.4 & 56.3 & 48.9 & \textbf{93.5} & 57.3 & 30.2 & 67.9 & 58.6 \\
   $\hookrightarrow$ w/ Scaffold-Stripping Mechanism  & 77.6 & 35.2 & 58.6 & 44.2 & 92.9 & 56.0 & \textbf{30.6} & \textbf{68.4} & 57.9 \\
    
    \rowcolor{mygray}
    \textbf{AnE-$1^{st}$(Ours)} & \textbf{79.6} & \textbf{39.7} & \textbf{60.5} & \textbf{49.1} & \textbf{93.5} & \textbf{57.7} & 30.5 & 68.3 & \textbf{59.9} \\
    \bottomrule
  \end{tabular}%
  }
\end{table*}

\paragraph{Effectiveness of Truth Anchor Expansion.} 
To verify the necessity of Truth Anchor Expansion in objective reality, we compare our Truth Anchor Expansion with random and synthetic alternatives (\cref{tab:ablation-expansion}). 
We compare Random Expansion and Synthetic Expansion, where Random Expansion randomly samples data from the external pool, and Synthetic Expansion generates data using Qwen2.5-VL-72B. All methods use the same data volume for a fair comparison.
As shown in \cref{tab:ablation-expansion}, compared to No Expansion, Random Expansion significantly reduces performance by 3.8\%, indicating that blind data augmentation may introduce a large amount of irrelevant data, which harms the reasoning abilities of multimodal models.
Compared to No Expansion, Synthetic Expansion yields comparable but slightly lower performance, suggesting that purely synthetic data may introduce noisy or hallucinated samples that offset its potential benefits. In contrast, Truth Anchor Expansion retrieves real samples around the failing frontier and consistently improves multimodal reasoning performance across most metrics.
\begin{table*}[t]
  \centering
  \small
  \caption{
    \textbf{Analysis of Expansion Strategies}, demonstrating that our Truth Anchor Expansion achieves faithful data curation to enhance model reasoning capabilities.
  }
  \label{tab:ablation-expansion}
  \renewcommand{\arraystretch}{1.2}
  \resizebox{\textwidth}{!}{%
  \begin{tabular}{l|cccccccc|c}
    \toprule
    \textbf{Setting} & \textbf{MathVista} & \textbf{MathVision} & \textbf{MathVerse} & \textbf{LogicVista} & \textbf{SciQA} & \textbf{MMMU} & \textbf{EMMA} & \textbf{MMSTAR} & \textbf{Avg} \\
    \midrule
    Preliminary RL & 77.9 & 32.8 & 53.6 & 45.5 & 92.6 & 57.7 & 30.1 & 67.2 & 57.2 \\
    No Expansion    & 77.6 & 35.2 & 58.6 & 44.2 & 92.9 & 56.0 & \textbf{30.6} & \textbf{68.4} & 57.9 \\
    Random Expansion & 75.4 & 29.6 & 49.3 & 42.2 & 90.8 & 52.8 & 26.2 & 66.6 & 54.1 \\
    Synthetic Expansion(Self-evolution) & 78.0 & 34.7 & 56.2 & 46.7 & 93.3 & 55.1 & 30.4 & 67.9 & 57.8 \\
    \rowcolor{mygray}
    \textbf{Truth Anchor Expansion(Ours)}& \textbf{79.6} & \textbf{39.7} & \textbf{60.5} & \textbf{49.1} & \textbf{93.5} & \textbf{57.7} & 30.5 & 68.3 & \textbf{59.9} \\
    \bottomrule
  \end{tabular}%
  }
\end{table*}

\begin{wrapfigure}{h}{2.3in}
\centering
\resizebox{2.3in}{!}{    
\includegraphics[width=0.6\linewidth]{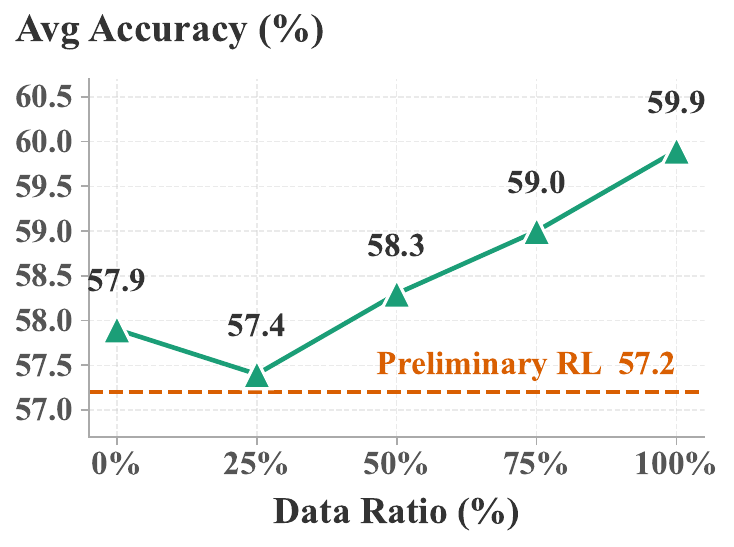}}
    \caption{\textbf{Data scaling law of Truth Anchor Expansion.} Average accuracy on eight benchmarks under different mixing ratios of Truth Anchor Expansion data. Performance improves as more anchor data is incorporated.}
\label{fig:data-scaling}
\end{wrapfigure}

\paragraph{Data Scaling Law of Truth Anchor Expansion.} 
Furthermore, we investigate the data scaling properties of Truth Anchor Expansion. 
As shown in \cref{fig:data-scaling}, we vary the proportion of Truth Anchor Expansion data mixed 
into the SFT training set to examine the effect of this augmented data. 
We observe that incorporating only a small fraction of the expanded data results in a slight performance dip, likely due to the distribution shift caused by the limited additional samples. 
However, as the mixing ratio increases, model performance improves steadily and monotonically, ultimately reaching an average accuracy of 59.9\% when the data is fully incorporated. 
This trend suggests that the data produced by Truth Anchor Expansion is scalable. Its benefits grow with volume, indicating that the retrieved samples offer genuinely diverse and informative training signals, rather than introducing redundant information.

\paragraph{Ablation Study on Scaffold-Stripping.}  
As shown in \cref{tab:ablation-internalization}, we perform an ablation study to assess the impact of different Scaffold-Stripping components. In the second row, directly distilling using the Teacher model's data results in a 2.4\% performance drop compared to using our Scaffold-Stripping method. This suggests that the reasoning trajectories from the Teacher model are difficult to internalize, causing the model to perform only shallow imitation instead of genuinely improving its reasoning ability.
In the fourth row, when the RL Stripping phase is removed and only the Scaffold-Augmented SFT phase is performed, there is a 1.5\% performance decrease. This indicates that the model cannot fully internalize the Teacher's guidance through the Scaffold-Augmented SFT phase alone.
Finally, as shown in the fifth row, if RL data does not include the \( R_{\text{frontier}}^{+} \) data from the SFT phase and RL is applied directly, the performance drops by 2.0\%. This highlights that the improvement in RL performance depends significantly on internalizing the teacher’s guidance from \( R_{\text{frontier}}^{+} \).

\begin{table*}[t]
  \centering
  \small
  \caption{
    \textbf{Impact of the Scaffold-Stripping Mechanism.}
Directly distilling from the teacher model yields only a marginal improvement over the baseline.
In contrast, removing the RL stripping stage, or running RL without scaffold-augmented data (i.e., without the hint-based scaffold), both cause a significant drop in AnE performance.
  }
  \label{tab:ablation-internalization}
  \renewcommand{\arraystretch}{1.2}
  \resizebox{\textwidth}{!}{%
  \begin{tabular}{l|cccccccc|c}
    \toprule
    \textbf{Setting} & \textbf{MathVista} & \textbf{MathVision} & \textbf{MathVerse} & \textbf{LogicVista} & \textbf{SciQA} & \textbf{MMMU} & \textbf{EMMA} & \textbf{MMSTAR} & \textbf{Avg} \\
    \midrule

        Preliminary RL & 77.9 & 32.8 & 53.6 & 45.5 & 92.6 & 57.7 & 30.1 & 67.2 & 57.2 \\
        Teacher Distillation & 77.8 & 34.7 & 51.3 & 47.1 & 92.9 & 57.0 & \textbf{31.4} & 67.8 & 57.5\\ 
    \rowcolor{mygray}
    
    \textbf{Scaffold-Stripping(Ours)} & \textbf{79.6} & \textbf{39.7} & \textbf{60.5} & \textbf{49.1} & 93.5 & 57.7 & 30.5 & \textbf{68.3} & \textbf{59.9} \\

    $\hookrightarrow$ w/o RL Stripping & 78.4 & 35.6 & 56.0 & 48.0 & \textbf{93.6} & \textbf{57.8} & 30.2 & 67.9 & 58.4 \\
    $\hookrightarrow$ w/o \( R_{\text{frontier}}^{+} \) in RL & 78.1 & 34.6 & 56.3 & 47.1 & 92.9 & 55.3 & 31.2 & 67.8 & 57.9 \\
      \bottomrule
  \end{tabular}%
  }
\end{table*}

\subsection{Case Study}
\begin{figure}

    \centering
    \includegraphics[width=1\linewidth]{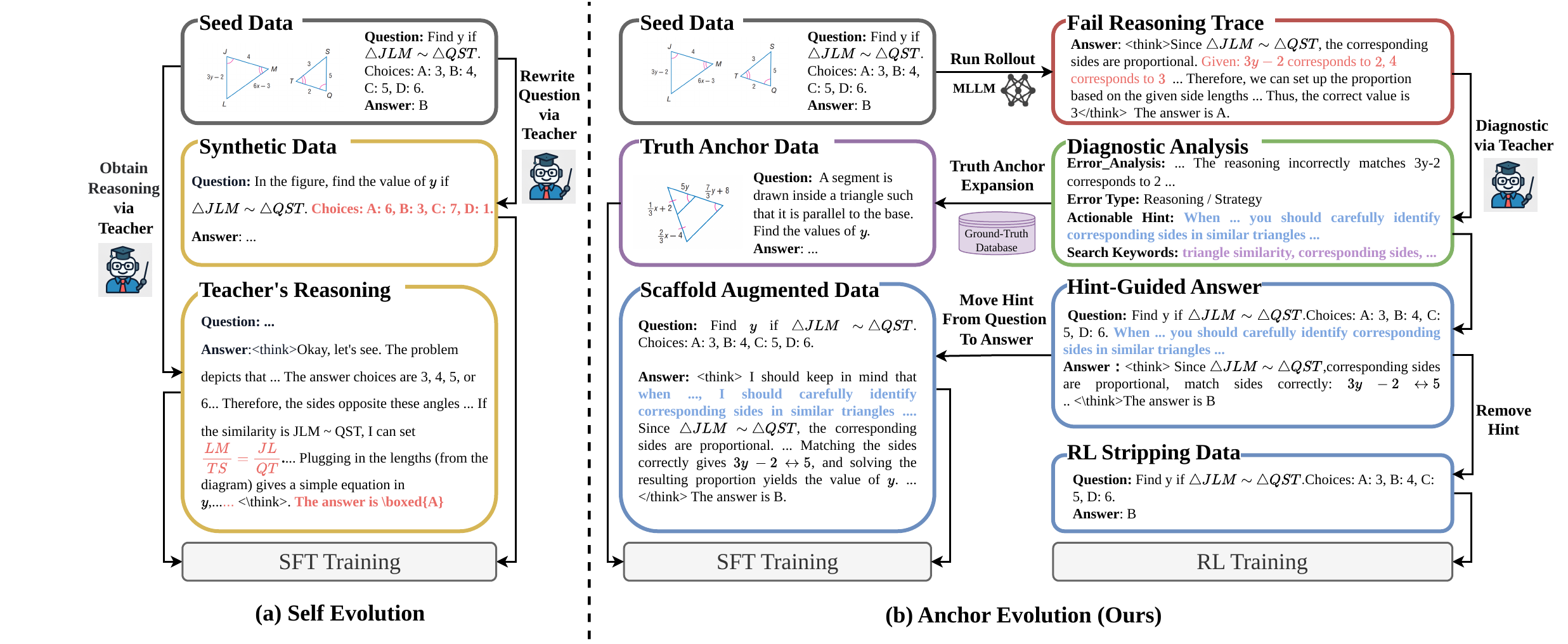} 
\caption{\textbf{Illustrative Examples.} 
\textbf{(a)} Prior self-evolution methods may synthesize low-quality data and generate overly complex, hallucination-prone reasoning that exceeds the model's current capabilities, limiting effective learning. 
\textbf{(b)} By analyzing failed reasoning, Anchor Evolution retrieves real data using search keywords and guides the model with actionable hints, helping it improve reasoning and develop independent problem-solving skills.}
    \label{fig:case}
\end{figure}

\cref{fig:case} shows a geometry example to contrast AnE with prior self-evolution methods. In \cref{fig:case}(a), teacher-generated rewrites and reasoning trajectories may contain hallucinated content (e.g., options missing the correct answer) and produce complex and hallucination-prone paths that are misaligned with the student model’s capability distribution, which can lead to superficial imitation and accumulated drift after training.
In \cref{fig:case}(b), AnE first collects the model’s failed trajectories via rollouts to locate the failing frontier. 
The teacher then extracts failure-relevant keywords and actionable hints.
The keywords trigger Truth Anchor Expansion to retrieve verified truth anchors from ground-truth databases, yielding valid, high-fidelity problems that target the model's weakness without hallucination. The hints are used for scaffold-augmented supervision, and the Scaffold-Stripping Mechanism further applies RL to strip the scaffold template, enabling the model to internalize the reasoning capability.
Overall, compared with prior self-evolution that relies on synthetic data and may accumulate hallucinated reasoning paths and cognitive drift, AnE anchors data evolution with verified truth anchors (Truth Anchor Expansion) and internalizes reasoning via the Scaffold-Stripping Mechanism, enabling stable improvements targeted at the failing frontier.

\section{Conclusion}
In this paper, we presented \textit{Anchor Evolution} (AnE), a post-training paradigm that integrates truth-anchored data curation and model evolution to achieve faithful and sustained gains at the multimodal reasoning frontier.
We showed that static post-training with static datasets is limited by non-extensible data distributions, while self-evolution based on synthetic data can suffer from cognitive drift and hallucinated reasoning paths.
To address these limitations, AnE identifies the model's failing frontier via trajectory rollouts and performs Truth Anchor Expansion by retrieving verified truth anchors from ground-truth databases, enabling high-fidelity data curation grounded in real-world knowledge.
Moreover, the proposed Scaffold-Stripping Mechanism first anchors reasoning paths through scaffold-augmented supervision and then leverages RL to strip the scaffold template, effectively internalizing reasoning capabilities.
Extensive experiments across multiple multimodal reasoning benchmarks demonstrate consistent improvements over the base model and state-of-the-art performance on challenging datasets, validating the effectiveness of AnE for faithful model evolution.

\bibliographystyle{unsrt} 
\bibliography{main}
\appendix

\renewcommand{\thesection}{\Alph{section}}
\counterwithin{table}{section}
\counterwithin{figure}{section}
\renewcommand{\thetable}{\thesection\arabic{table}}
\renewcommand{\thefigure}{\thesection\arabic{figure}}

\tcbset{
  appendixpromptcompact/.style={
    enhanced,
    width=\linewidth,
    colback=white,
    colframe=black,
    boxrule=0.8pt,
    arc=1.5mm,
    left=1mm,
    right=1mm,
    top=1mm,
    bottom=1mm,
    attach boxed title to top left={xshift=2mm,yshift=-1.5mm},
    boxed title style={
      colback=black,
      colframe=black,
      boxrule=0pt,
      arc=0.8mm
    },
    fonttitle=\bfseries\footnotesize,
    coltitle=white,
    listing only,
listing options={
  basicstyle=\ttfamily\scriptsize,
  breaklines=true,
  breakatwhitespace=false,
  columns=fullflexible,
  keepspaces=true,
  breakautoindent=false,
  breakindent=0pt
}
  }
}

\tcbset{
  failureframe/.style={
    enhanced,
    width=\linewidth,
    colback=white,
    colframe=black,
    boxrule=0.85pt,
    arc=1.6mm,
    left=2mm,
    right=2mm,
    top=2mm,
    bottom=2mm,
    attach boxed title to top left={xshift=2.5mm,yshift=-1.8mm},
    boxed title style={
      colback=black,
      colframe=black,
      boxrule=0pt,
      arc=1mm
    },
    fonttitle=\bfseries\footnotesize
  },
  innersubbox/.style={
    colback=white,
    colframe=black!35,
    boxrule=0.45pt,
    arc=1mm,
    left=1.2mm,
    right=1.2mm,
    top=0.9mm,
    bottom=0.9mm,
    fonttitle=\bfseries\footnotesize,
    fontupper=\scriptsize
  },
  infocard/.style={
    colback=gray!5,
    colframe=black!30,
    boxrule=0.4pt,
    arc=0.9mm,
    left=1mm,
    right=1mm,
    top=0.8mm,
    bottom=0.8mm,
    fontupper=\scriptsize
  },
  responsecard/.style={
    colback=gray!3,
    colframe=black!30,
    boxrule=0.4pt,
    arc=0.9mm,
    left=1mm,
    right=1mm,
    top=0.8mm,
    bottom=0.8mm,
    fonttitle=\bfseries\footnotesize,
    fontupper=\scriptsize
  },
  analysiscard/.style={
    colback=gray!4,
    colframe=black!20,
    boxrule=0.35pt,
    arc=0.8mm,
    left=0.9mm,
    right=0.9mm,
    top=0.7mm,
    bottom=0.7mm,
    fonttitle=\bfseries\footnotesize,
    fontupper=\scriptsize
  }
}
\tcbset{
  failureframe/.style={
    enhanced,
    width=\linewidth,
    colback=white,
    colframe=black,
    boxrule=0.8pt,
    arc=1.4mm,
    left=1.8mm,
    right=1.8mm,
    top=1.8mm,
    bottom=1.8mm,
    attach boxed title to top left={xshift=2.2mm,yshift=-1.5mm},
    boxed title style={
      colback=black,
      colframe=black,
      boxrule=0pt,
      arc=0.9mm
    },
    fonttitle=\bfseries\footnotesize
  },
  innersubbox/.style={
    colback=white,
    colframe=black!35,
    boxrule=0.45pt,
    arc=0.9mm,
    left=1.0mm,
    right=1.0mm,
    top=0.8mm,
    bottom=0.8mm,
    fonttitle=\bfseries\footnotesize,
    fontupper=\scriptsize
  },
  infocard/.style={
    colback=gray!5,
    colframe=black!30,
    boxrule=0.35pt,
    arc=0.7mm,
    left=0.7mm,
    right=0.7mm,
    top=0.45mm,
    bottom=0.45mm,
    fontupper=\tiny
  },
  responsecard/.style={
    colback=gray!3,
    colframe=black!30,
    boxrule=0.35pt,
    arc=0.7mm,
    left=0.8mm,
    right=0.8mm,
    top=0.6mm,
    bottom=0.6mm,
    fonttitle=\bfseries\footnotesize,
    fontupper=\scriptsize
  },
  analysiscard/.style={
    colback=gray!4,
    colframe=black!20,
    boxrule=0.35pt,
    arc=0.8mm,
    left=0.9mm,
    right=0.9mm,
    top=0.7mm,
    bottom=0.7mm,
    fonttitle=\bfseries\footnotesize
  }
}

\clearpage
\begin{center}
    {\Large \bfseries Supplementary Material}
\end{center}

\section{Details of Experimental Settings}
\label{sec:appendix-settings}
\paragraph{Experimental Settings of Failing-Frontier Discovery.}
At this stage, Qwen2.5-VL-72B-Instruct is used as the teacher model for diagnostic analysis of failure samples.
\cref{tab:teacher-prompt} shows the prompt used for the teacher model in Failing-Frontier Discovery. The teacher output is formatted as structured JSON, from which \texttt{search\_keywords} and \texttt{actionable\_hint} are extracted using regular-expression matching. These two fields are then used in the subsequent stages of Truth Anchor Expansion and the Scaffold-Stripping Mechanism.
\begin{table*}[t]
    \centering
        \caption{\textbf{Prompt used for the teacher model.} System prompt used to diagnose model failures and generate structured feedback.}
    \begin{tcblisting}{
        appendixpromptcompact,
        title={Prompt used for the Teacher Model}
    }
You are an expert AI QA specialist. Analyze an incorrect vision-language model response and identify the primary and secondary causes of failure.
Input
Question: {question}
Ground Truth: {ground_truth}
Model Response: {response}
Instructions
1. Failure Analysis
- Locate the exact step where the logic first deviates.
- Quote the erroneous or hallucinated sentence(s).
- Contrast with reality using statements such as:
  "Model claimed [X], but the image/text shows [Y]."
  "Model calculated [X], but the correct derivation is [Y]."
2. Error Categorization
Choose exactly one primary root cause. Add secondary factors only if necessary. If the primary cause fully explains the failure, leave secondary_errors empty.
Categories
- Perception / Understanding: incorrect reading of the input.
- Knowledge / Information: correct perception but missing or wrong external knowledge.
- Reasoning / Strategy: correct perception and knowledge, but incorrect processing, logic, or calculation.
- Execution / Output: correct reasoning but incorrect final formatting or output.
3. Feedback
- Explanation: concise diagnosis of why the failure happened.
- Actionable Hint: must contain both:
  Strategy: high-level advice.
  Mechanism: trigger -> action -> verification.
4. Search Keywords
Provide 3-6 short English phrases for retrieving similar practice examples.
Use phrases likely to appear in similar questions.
Keyword focus:
- Perception / Understanding: visual task description or input format.
- Knowledge / Information: specific concept or fact.
- Reasoning / Strategy: problem structure or logical step.
- Execution / Output: required formatting instruction.
Output
Return one valid JSON object inside a list. Do not use markdown code blocks.
[{
  "error_analysis": "Step-by-step diagnosis. Quote the error. Contrast with reality.",
  "primary_error": {
    "error_type": "Category",
    "explanation": "Root cause explanation.",
    "actionable_hint": "Strategy: [Macro]. Mechanism: [Micro].",
    "search_keywords": ["phrase 1", "phrase 2"]
  },
  "secondary_errors": [
    {
      "error_type": "Category",
      "explanation": "Explanation.",
      "actionable_hint": "Strategy: [Macro]. Mechanism: [Micro].",
      "search_keywords": ["keyword"]
    }
  ]
}]
    \end{tcblisting}

    \label{tab:teacher-prompt}
\end{table*}

\paragraph{Experimental Settings of Truth Anchor Expansion.}
Before Anchor Evolution training, all samples in the external ground-truth database FineVision are encoded into dense embeddings using Qwen3-Embedding-8B. 
This database-side embedding is performed only once and takes approximately 8 hours on 64 Ascend 910B NPUs. 
During Truth Anchor Expansion, for each query sample, the extracted \texttt{search\_keywords} are first normalized with the prompt shown in \cref{tab:appendix-embedding-prompt} and then encoded by Qwen3-Embedding-8B. 
Based on embedding similarity, the top-\(N\) most similar samples are retrieved to form the raw anchor set \(\mathcal{A}_{\text{raw}}\). After deduplication and quality filtering, the final anchor set \(\mathcal{A}\) is obtained and used to expand the SFT dataset.

\begin{table*}[t]
    \centering
        \caption{\textbf{Prompt used for the embedding model in Truth Anchor Expansion.} 
    It is used to convert \texttt{search\_keywords} into retrieval queries for embedding-based similarity search.}
\begin{tcblisting}{
    appendixpromptcompact,
    title={Prompt used for the Embedding Model}
}
You are retrieving solved vision-language QA examples from a high-quality training database. The query provides search keywords, phrases, or specific terminology derived from a failure labeled as {error_type}. Retrieve questions that contain similar language or concepts, matching the task type, input representation (e.g., text, tables, charts, or documents), and reasoning structure described by these keywords. The goal is to find correct examples that serve as truth anchors for this specific failure mode.
Query: {search_keywords}
\end{tcblisting}

    \label{tab:appendix-embedding-prompt}
\end{table*}
\paragraph{Experimental Settings of the Scaffold-Stripping Mechanism.}
In this stage, the extracted \texttt{actionable\_hint} is appended to the original question with a randomly selected teacher-style template to guide the model to generate a correct response in the guided second rollout.
For each successful guided trajectory, the hint is then moved from the question side to the response side and reformatted with a randomly selected student-style template. The resulting samples are used to construct scaffold-augmented SFT data for subsequent SFT training.

\paragraph{Experimental Settings of SFT and RL.}
SFT is implemented with MindSpeed-MM (v2.3.0) on 64 Ascend 910B NPUs and trained until performance converges on the validation set. AdamW is used as the optimizer with a cosine learning-rate schedule, a learning rate of \(5.0\times10^{-6}\), zero weight decay, dynamic batch sizing, and a maximum sequence length of 8192. To reduce the memory overhead of long sequences, non-uniform sequence parallelism and gradient checkpointing are enabled. RL is conducted with veRL (v0.7.0) on 64 Ascend 910B NPUs following the GSPO training recipe, with AdamW as the optimizer, a constant learning-rate schedule, a learning rate of \(1.0\times10^{-6}\), weight decay of 0.1, dynamic batch sizing, a maximum sequence length of 16384, and 16 rollouts enabled. RL training is also continued until performance converges on the validation set.

\paragraph{Hallucination Evaluation Details.}
We evaluate hallucinated reasoning in the constructed post-training data with an image-aware LLM-as-judge protocol.
The evaluation covers two data sources: \textit{Self-Evolution} samples and \textit{Truth Anchor} samples retrieved by AnE from FineVision.
For each source, we uniformly sample \(N\) examples, with \(N{=}1000\), and use \texttt{gpt-4o}~\cite{hurst2024gpt} as the judge.
Given one or multiple images, the question, and the assistant answer, the judge assigns a single hallucination score from 0 to 5, where higher scores indicate more severe hallucination or task deviation.
We define samples with a hallucination score \(\geq 4\) as hallucinated samples and report the corresponding hallucination rate.
The judge also assigns a hallucination-type label and provides a one-sentence explanation.
The detailed judge prompt is provided in \cref{tab:appendix-hallucination-prompt}.
\begin{table*}[t]
    \centering
    \caption{\textbf{User prompt for hallucination evaluation.}
    The judge inspects one or multiple images, the question, and the answer, then assigns a single hallucination score and type label.}
    \begin{tcblisting}{
        appendixpromptcompact,
        title={User Prompt for Hallucination Evaluation}
    }
You are given a vision-language Q&A sample, including one or multiple images, a question, and an assistant answer.

[IMAGE(S)]
{images}

[QUESTION]
{question}

[ASSISTANT ANSWER]
{answer}

Evaluate whether the assistant answer contains hallucinated reasoning, unsupported visual/textual claims, or task deviation according to the image(s) and the question.

Assign one integer HALLUCINATION score from 0 to 5, where higher means worse:

HALLUCINATION 0-5
- 0: No hallucination. The answer is visually grounded, on task, and logically consistent.
- 1: Very minor imprecision or irrelevant detail, with no effect on the reasoning or final answer.
- 2: Minor unsupported or weakly grounded claim, but the answer is still mostly correct and on task.
- 3: Clear unsupported visual/textual claim, reasoning gap, or partial task deviation that affects the reasoning.
- 4: Severe hallucination. The answer contains major fabricated, image-inconsistent, logically contradictory, or substantially off-task content.
- 5: Dominant hallucination. The answer is completely wrong, fabricated, image-inconsistent, or answers a different task.

HALLUCINATION_TYPE in:
"none", "visual_error", "factual_error", "fabricated_entity", "contradiction", "unsupported_reasoning", "task_drift"

Return STRICT JSON only:
{"hallucination": 0, "type": "none", "explanation": "The answer is visually grounded, on task, and logically consistent."}
    \end{tcblisting}
    \label{tab:appendix-hallucination-prompt}
\end{table*}
\paragraph{Evaluation Details}
For the reproduced methods in Table~1, the evaluation settings follow those reported in the corresponding papers as closely as possible. For our method, a unified evaluation protocol is adopted for both SFT and RL models. During inference, a predefined system prompt (shown in \cref{tab:appendix-eval-prompt}) is applied to enforce a structured output format, requiring the model to generate both a complete reasoning trace and a clearly identifiable final answer for automated evaluation.
Model predictions are assessed through a hierarchical validation pipeline for answer extraction and correctness verification. A rule-based validator is first applied for answer extraction and preliminary checking to minimize computational cost. When the rule-based mechanism fails to reliably extract or verify the answer, an LLM-as-judge module is further invoked for supplementary adjudication.
vLLM is employed as the serving engine to accelerate inference. 

\begin{table*}[t]
    \centering
        \caption{\textbf{Prompt used for evaluation.} 
    The prompt enforces a structured output format, including an explicit reasoning trace and a boxed final answer.}
    \begin{tcblisting}{
        appendixpromptcompact,
        title={Prompt used for Evaluation}
    }
{question} You FIRST think about the reasoning process as an internal monologue and then provide the final answer. The reasoning process MUST BE enclosed within <think> </think> tags. The final answer MUST BE put in  \\boxed{}.
    \end{tcblisting}

    \label{tab:appendix-eval-prompt}
\end{table*}

\section{Computational Cost of AnE}
\label{sec:appendix-cost}
\cref{tab:ane-cost} reports the computational cost of each module in Anchor Evolution, taking the first evolution round as an example. In later rounds, the time consumption of each module decreases accordingly because the SFT and RL datasets become smaller. As a result, the total computational cost is reduced from 76.2 hours in Round 1 to 61.5 hours in Round 2 and 53.3 hours in Round 3. Overall, Truth Anchor Expansion introduces negligible overhead, while the major computational cost mainly comes from teacher-based diagnostic analysis in Failing-Frontier Discovery and RL training in the Scaffold-Stripping Mechanism.
\begin{table}[t]
  \centering
  \small
\caption{\textbf{Computational cost of Anchor Evolution in Round 1.} }
  \label{tab:ane-cost}
  \renewcommand{\arraystretch}{1.12}
  \resizebox{0.7\linewidth}{!}{%
  \begin{tabular}{llcc}
    \toprule
    \textbf{Module} & \textbf{Subcomponent} & \textbf{Time (h)} & \textbf{Ratio (\%)} \\
    \midrule
    \multirow{3}{*}{Failing-Frontier Discovery}
      & \(K\)-rollout on seed data & \(\sim\)10.0 & 13.1 \\
      & Teacher-based diagnostic analysis & \(\sim\)22.0 & 28.9 \\
      & \textbf{Module Total} & \textbf{\(\sim\)32.0} & \textbf{42.0} \\
    \midrule
    \multirow{3}{*}{Truth Anchor Expansion}
      & Embedding of retrieval queries & \(\sim\)0.1 & 0.1 \\
      & Similarity-based retrieval & \(\sim\)0.1 & 0.1 \\
      & \textbf{Module Total} & \textbf{\(\sim\)0.2} & \textbf{0.3} \\
    \midrule
    \multirow{4}{*}{Scaffold-Stripping Mechanism}
      & Hint-guided second rollout & \(\sim\)5.0 & 6.6 \\
      & Scaffold-Augmented SFT & \(\sim\)8.0 & 10.5 \\
      & RL stripping training & \(\sim\)31.0 & 40.7 \\
      & \textbf{Module Total} & \textbf{\(\sim\)44.0} & \textbf{57.7} \\
    \midrule
    \multicolumn{2}{l}{\textbf{Total}} & \textbf{\(\sim\)76.2} & \textbf{100.0} \\
    \bottomrule
  \end{tabular}%
  }
\end{table}

In addition to the module-wise cost breakdown, we further compare AnE with simpler post-training baselines under a similar compute budget.
As shown in \cref{tab:cost-performance}, AnE-$1^{st}$ (including the cost of Preliminary RL) achieves 59.9\% average accuracy with 105.5h training time.
It outperforms RL-only by +2.4 points (57.5\%$\rightarrow$59.9\%) and one-pass SFT+RL by +1.7 points (58.2\%$\rightarrow$59.9\%), while using 8.0h and 7.2h less training time, respectively.
This suggests that AnE provides a better cost-performance trade-off, rather than simply relying on additional computation.

\begin{table*}[t]
  \centering
  \small
  \caption{
  \textbf{Cost-performance comparison under a similar compute budget.}
  The cost of AnE-$1^{st}$ includes Preliminary RL.
  AnE achieves better performance than RL-only and one-pass SFT+RL with comparable overall training time.
  }
  \label{tab:cost-performance}
  \renewcommand{\arraystretch}{1.15}
  \resizebox{\textwidth}{!}{%
  \begin{tabular}{l|c|cccccccc|c}
    \toprule
    \textbf{Setting} & \textbf{Time} & \textbf{MathVista} & \textbf{MathVision} & \textbf{MathVerse} & \textbf{LogicVista} & \textbf{SciQA} & \textbf{MMMU} & \textbf{EMMA} & \textbf{MMSTAR} & \textbf{Avg} \\
    \midrule
    RL-only 
    & 113.5h & 77.6 & 33.7 & 51.3 & 48.2 & 92.3 & 57.7 & \textbf{31.1} & \textbf{68.3} & 57.5 \\
    
    One-pass SFT+RL 
    & 112.7h & 78.1 & 34.0 & 58.2 & 47.8 & 92.5 & \textbf{58.8} & 29.9 & 66.6 & 58.2 \\
    
    \textbf{AnE-$1^{st}$} 
    & \textbf{105.5h} & \textbf{79.6} & \textbf{39.7} & \textbf{60.5} & \textbf{49.1} & \textbf{93.5} & 57.7 & 30.5 & \textbf{68.3} & \textbf{59.9} \\
    \bottomrule
  \end{tabular}%
  }
\end{table*}

\section{Training Curves of RL}
\label{sec:appendix-training-curves}
\begin{figure*}[t]
\centering
\includegraphics[width=\textwidth]{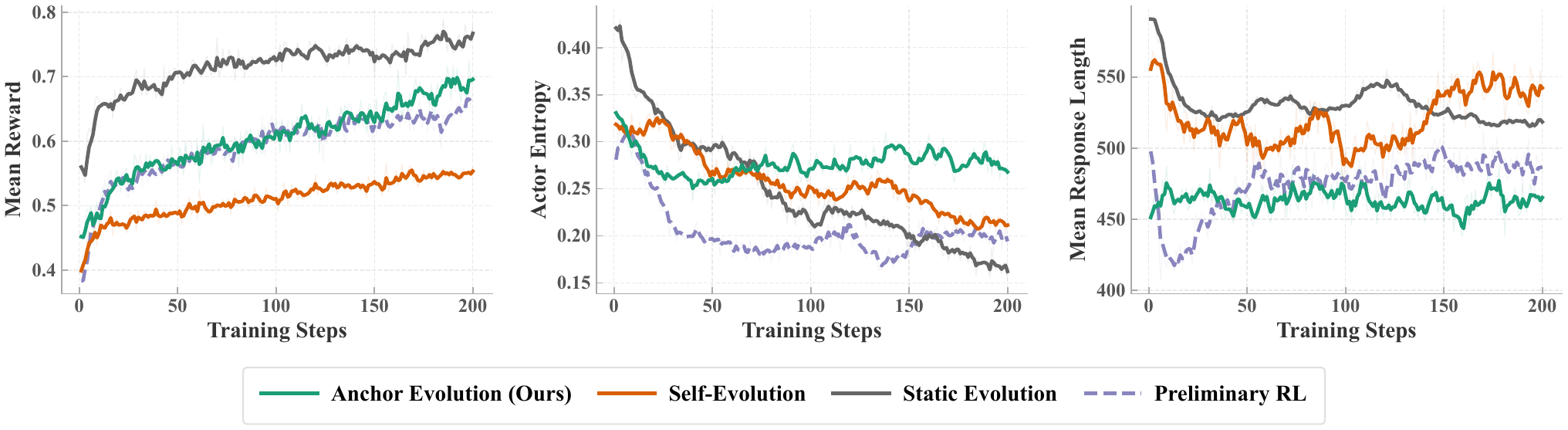}
\caption{\textbf{Training dynamics under different post-training paradigms.}
Comparison of AnE, Self-Evolution, Static Evolution, and the Preliminary RL baseline in terms of mean reward, actor entropy, and mean response length during RL training.}
\label{fig:training-curves}
\end{figure*}

As shown in \cref{fig:training-curves}, we compare the RL training curves of three methods in Round 1, along with the Preliminary RL baseline. For the reward curve, Anchor Evolution continues to improve beyond Preliminary RL, indicating that the challenging scaffold-free samples introduced from the failing frontier provide additional learning signals for RL. 
In contrast, Static Evolution quickly saturates because it uses the same static RL data as Preliminary RL, while Self-Evolution shows slower reward growth, likely due to noisy or hallucinated supervision introduced in its SFT stage. For the entropy curve, Anchor Evolution maintains higher policy entropy than Self-Evolution and Static Evolution, suggesting stronger exploration throughout RL training. For the response-length curve, Anchor Evolution produces shorter responses than methods relying on longer teacher-generated or manually annotated reasoning traces, indicating higher token efficiency. This is likely because its concise \texttt{actionable\_hint} guides the model toward successful reasoning trajectories within its own capability distribution. Therefore, it achieves higher token efficiency and avoids unnecessarily verbose responses, while still improving downstream reasoning performance.
\section{Data Size over Rounds}
\label{sec:appendix-data-size}

\begin{wrapfigure}{r}{0.52\linewidth}
    \centering
    \vspace{-6mm}
    \includegraphics[width=\linewidth]{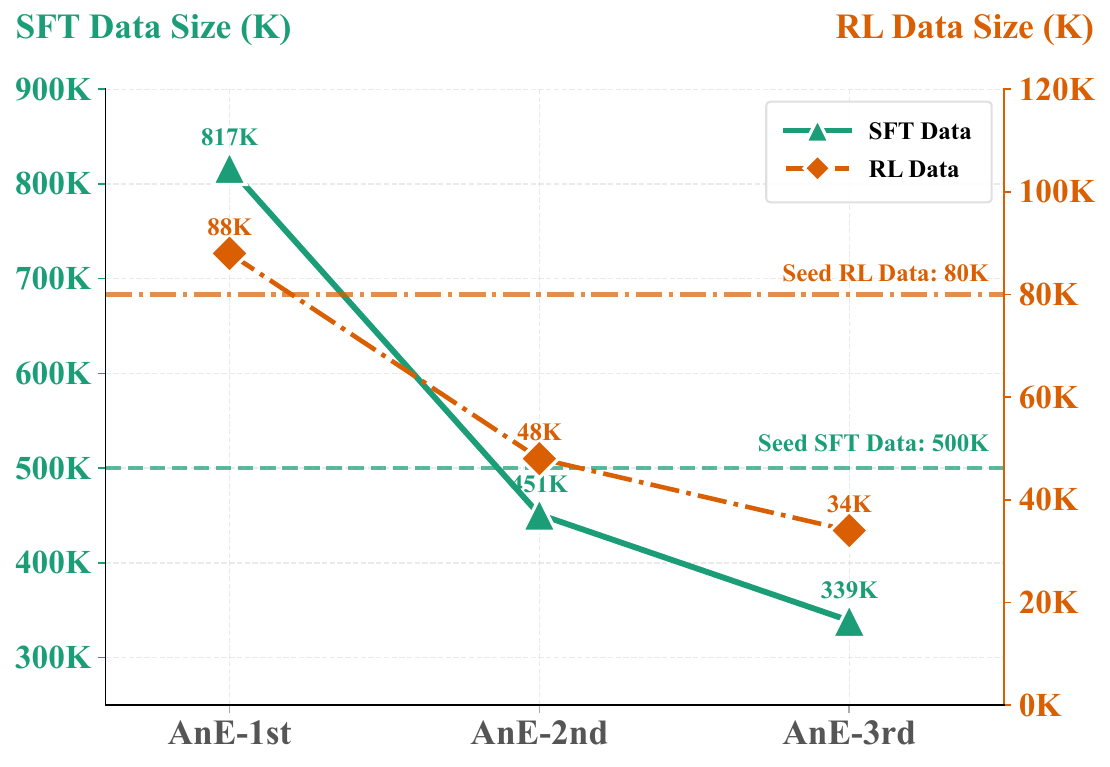}
    \caption{\textbf{SFT and RL data size across evolution rounds.} 
    Size of the SFT and RL datasets in AnE-$1^{st}$, AnE-$2^{nd}$, and AnE-$3^{rd}$, compared with the corresponding seed datasets.}
    \label{fig:data-size-rounds}
    \vspace{-3mm}
\end{wrapfigure}

\cref{fig:data-size-rounds} shows the size of the SFT and RL datasets across different evolution 
rounds. Compared with the seed datasets, AnE-$1^{st}$ substantially expands both the SFT and RL data, reaching 817K and 88K samples, respectively. As the evolution proceeds, the data size gradually decreases to 451K/48K in AnE-$2^{nd}$ and 339K/34K in AnE-$3^{rd}$. This trend suggests that, after the first round, fewer new samples remain near the model's failing frontier, while the newly constructed data become increasingly concentrated on harder and more persistent failure cases.

\section{More Ablation Studies}
\label{sec:appendix-ablation}

\begin{figure}[t]
    \centering
    \begin{minipage}[t]{0.48\linewidth}
        \centering
        \includegraphics[width=\linewidth]{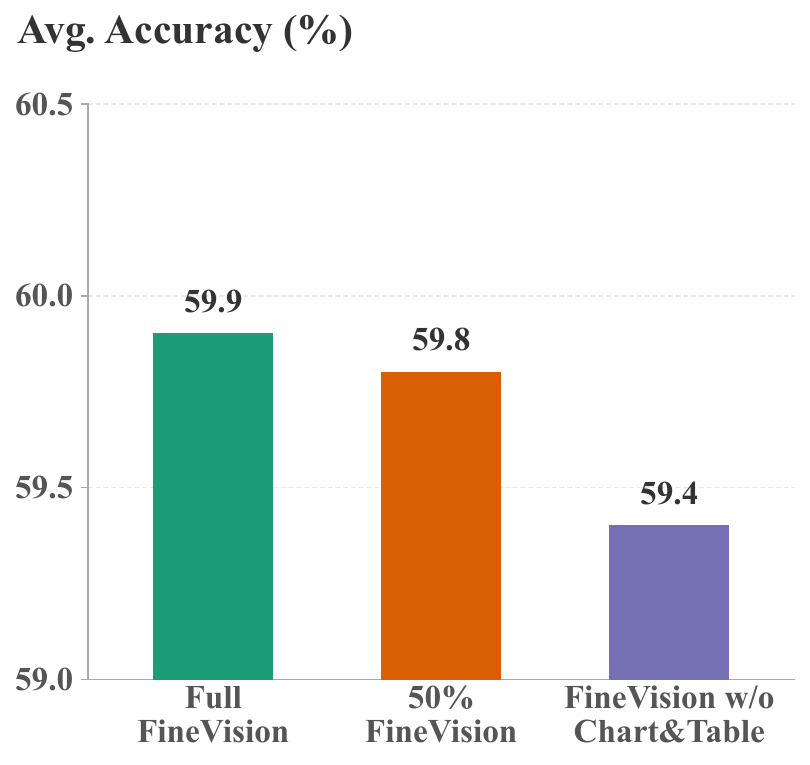}
        \vspace{-1mm}
        \centerline{\small (a) Avg. on eight benchmarks}
    \end{minipage}
    \hfill
    \begin{minipage}[t]{0.48\linewidth}
        \centering
        \includegraphics[width=\linewidth]{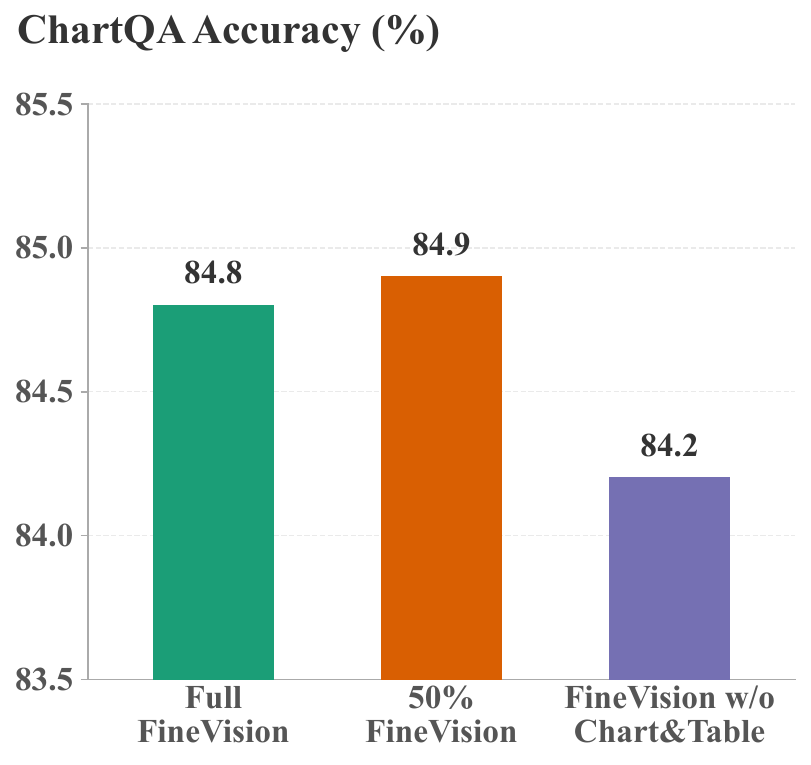}
        \vspace{-1mm}
        \centerline{\small (b) ChartQA}
    \end{minipage}
    \caption{
    \textbf{Robustness to database scale and domain coverage.}
    \textbf{(a)} Reducing FineVision to 50\% barely affects the average performance on the eight main benchmarks, while removing Chart\&Table samples causes a moderate drop.
    \textbf{(b)} ChartQA is reported as a domain-specific probe, where removing Chart\&Table samples slightly decreases performance.
    }
    \label{fig:database-coverage}
\end{figure}

\paragraph{Robustness to Database Scale and Domain Coverage.}
Since Truth Anchor Expansion retrieves anchors from an external ground-truth database, we further analyze the robustness of AnE to database scale and domain coverage.
We consider two settings: reducing FineVision to 50\% of its original size, and removing Chart\&Table-related samples from FineVision.
As shown in \cref{fig:database-coverage}(a), using only 50\% of FineVision barely affects the average performance over the eight main benchmarks, decreasing from 59.9\% to 59.8\%.
When Chart\&Table-related samples are removed, the average score decreases to 59.4\%, indicating that domain coverage affects retrieval effectiveness but does not cause AnE to collapse.
To further examine domain-specific coverage, we report ChartQA~\cite{masry2022chartqa} in \cref{fig:database-coverage}(b).
Removing Chart\&Table samples decreases ChartQA from 84.8\% to 84.2\%, while using 50\% FineVision maintains a comparable score of 84.9\%.
These results suggest that AnE is robust to moderate database reduction, while broad domain coverage remains beneficial for domain-specific reasoning tasks.
When anchor coverage is limited, Truth Anchor Expansion contributes less, but verified-anchor filtering and the Scaffold-Stripping Mechanism still provide effective learning signals.

\paragraph{Data Leakage Analysis.}
To further rule out potential data leakage, we measure the semantic overlap between retrieved truth anchors and evaluation samples.
Specifically, for each evaluation sample, we compute its nearest-neighbor similarity to the retrieved anchors from AnE-$1^{st}$ using an embedding model.
The average nearest-neighbor similarity is only 0.53, indicating low overall semantic overlap.
Moreover, fewer than 0.01\% of anchors exceed a similarity threshold of 0.95, corresponding to only 9 samples.
These results suggest that the retrieved anchors are unlikely to be near-duplicates of evaluation samples, and the performance gains of AnE are not caused by semantic leakage.

\paragraph{Comparison of Guidance Strategies.}
\cref{tab:answer-vs-hint} compares different guidance strategies, including answer-as-hint, teacher distillation, input-side hinting, and our hint-relocation strategy.
Although answer-as-hint provides direct target information, it mainly exposes the answer and brings limited improvement in intrinsic reasoning ability.
Teacher distillation also yields weaker performance, likely because teacher-generated responses are often verbose and mismatched with the student's capability distribution.
We further test \textit{Input-side Hinting}, where the hint is kept in the input during SFT and removed during RL. This variant obtains only 57.3\% average accuracy, suggesting that input-side hints may cause the model to rely on external guidance.
In contrast, our strategy relocates the teacher-generated \texttt{actionable\_hint} to the answer prefix during SFT and removes it during RL, thereby encouraging the model to internalize the reasoning strategy more effectively and achieving the best performance.
\begin{table*}[t]
  \centering
  \small
\caption{\textbf{Comparison of guidance strategies.}
Answer-as-hint, teacher distillation, and input-side hinting all provide weaker downstream gains, while our hint-relocation strategy achieves the best overall performance.}
  \label{tab:answer-vs-hint}
  \renewcommand{\arraystretch}{1.15}
  \resizebox{\textwidth}{!}{%
  \begin{tabular}{l|cccccccc|c}
    \toprule
    \textbf{Hint Type} & \textbf{MathVista} & \textbf{MathVision} & \textbf{MathVerse} & \textbf{LogicVista} & \textbf{SciQA} & \textbf{MMMU} & \textbf{EMMA} & \textbf{MMSTAR} & \textbf{Avg} \\
    \midrule
    Answer as Hint 
    & 76.9 & 34.1 & 53.3 & 47.8 & 92.9 & 53.1 & 31.2 & \textbf{68.4} & 57.2 \\
    
    Teacher Distillation 
    & 77.8 & 34.7 & 51.3 & 47.1 & 92.9 & 57.0 & 31.4 & 67.8 & 57.5 \\
    
Input-side Hinting & 77.9 & 34.7 & 54.9 & 46.7 & 91.8 & 55.7 & 30.4 & 66.1 & 57.3 \\
    
    \textbf{Ours} 
    & \textbf{79.6} & \textbf{39.7} & \textbf{60.5} & \textbf{49.1} & \textbf{93.5} & \textbf{57.7} & 30.5 & 68.3 & \textbf{59.9} \\
    \bottomrule
  \end{tabular}%
  }
\end{table*}

\paragraph{Comparison between search keywords and the original question in Truth Anchor Expansion.}
In Truth Anchor Expansion, retrieval can use either \texttt{search\_keywords} or the original question as the query. As shown in \cref{tab:keyword-vs-question}, directly using the original question tends to retrieve overly similar samples, emphasizing surface-level question similarity while failing to capture the underlying failure mode. As a result, the retrieved anchor data is less diverse and less targeted. In contrast, the \texttt{search\_keywords} focuses retrieval more directly on the failure pattern revealed by the incorrect response, leading to more diverse and more relevant truth anchors. \cref{tab:ablation-retrieval-query} further shows that using \texttt{search\_keywords} better performance than directly using the original question.

\begin{table*}[t]
    \centering
        \caption{\textbf{Example comparing retrieval with the original question and with \texttt{search\_keywords} in Truth Anchor Expansion.}
    Direct retrieval with the original question tends to return near-duplicate samples, whereas retrieval with \texttt{search\_keywords} better captures the underlying failure pattern and yields more diverse truth anchors.}
    \scriptsize
    \begin{tcolorbox}[
        failureframe,
        title=\textcolor{white}{\textbf{Comparison of Retrieval Queries in Truth Anchor Expansion}}
    ]

    \begin{tcolorbox}[infocard]
    \fontsize{6.8}{7.4}\selectfont
    \begin{minipage}[t]{0.24\linewidth}
        \vspace{0pt}
        \centering
        \includegraphics[width=\linewidth]{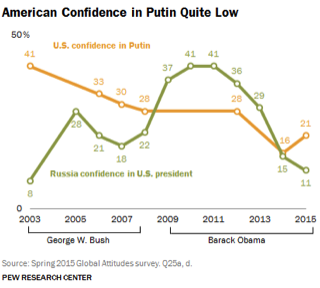}
    \end{minipage}
    \hfill
    \begin{minipage}[t]{0.73\linewidth}
        \vspace{0pt}
        \textbf{Question.} Which year did the green line cross the orange line from top to down?

        \vspace{2pt}
        \textbf{Answer.} \(\boxed{2011}\)
    \end{minipage}
    \end{tcolorbox}

    \vspace{3pt}

    \begin{tcolorbox}[
        analysiscard,
        title=\textbf{Retrieval with the Original Question},
        fonttitle=\bfseries
    ]
    \fontsize{6.8}{7.4}\selectfont
    \begin{minipage}[t]{0.24\linewidth}
        \vspace{0pt}
        \centering
        \includegraphics[width=\linewidth]{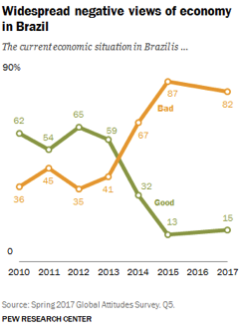}
    \end{minipage}
    \hfill
    \begin{minipage}[t]{0.73\linewidth}
        \vspace{0pt}
        \textbf{Retrieved Sample.} Which year did the orange line cross the green line?

        \vspace{2pt}
        \textbf{Answer.} \(\boxed{2013}\)
    \end{minipage}
    \end{tcolorbox}

    \vspace{3pt}

    \begin{tcolorbox}[
        analysiscard,
        title=\textbf{Retrieval with Search Keywords},
        fonttitle=\bfseries
    ]
    \fontsize{6.8}{7.4}\selectfont
    \begin{minipage}[t]{0.24\linewidth}
        \vspace{0pt}
        \centering
        \includegraphics[width=\linewidth]{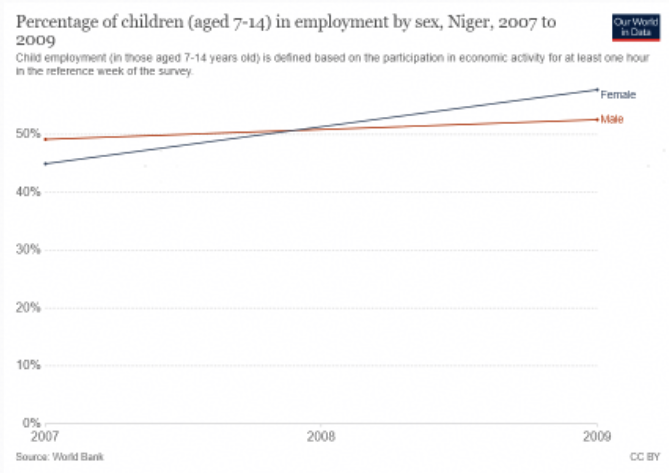}
    \end{minipage}
    \hfill
    \begin{minipage}[t]{0.73\linewidth}
        \vspace{0pt}
        \textbf{Search Keywords.} "line intersection on graph", "when two lines cross", "time-series line crossing"

        \vspace{2pt}
        \textbf{Retrieved Sample.} At how many points did the male and female lines in the given graph cross each other?

        \vspace{2pt}
        \textbf{Answer.} \(\boxed{1}\)
    \end{minipage}
    \end{tcolorbox}

    \end{tcolorbox}

    \label{tab:keyword-vs-question}
\end{table*}

\begin{table*}[t]
  \centering
  \small
  \caption{\textbf{Comparison between retrieval with the original question and retrieval with \texttt{search\_keywords} in Truth Anchor Expansion.} Retrieval with \texttt{search\_keywords} consistently achieves better performance than directly using the original question as the retrieval query.}
  \label{tab:ablation-retrieval-query}
  \renewcommand{\arraystretch}{1.15}
  \resizebox{\textwidth}{!}{%
  \begin{tabular}{l|cccccccc|c}
    \toprule
    \textbf{Setting} & \textbf{MathVista} & \textbf{MathVision} & \textbf{MathVerse} & \textbf{LogicVista} & \textbf{SciQA} & \textbf{MMMU} & \textbf{EMMA} & \textbf{MMSTAR} & \textbf{Avg} \\
    \midrule
    Retrieval with the Original Question & 78.2 & 38.1 & 58.2 & 47.8 & 91.7 & 55.1 & \textbf{30.9} & 68.0 & 58.5 \\

    \textbf{Retrieval with Search Keywords (Ours)} & \textbf{79.6} & \textbf{39.7} & \textbf{60.5} & \textbf{49.1} & \textbf{93.5} & \textbf{57.7} & 30.5 & \textbf{68.3} & \textbf{59.9} \\
    \bottomrule
  \end{tabular}
  }
\end{table*}

\paragraph{Comparison of Teacher Models in Diagnostic Analysis.}
\label{subsec:teacher-model}
We further
\begin{wrapfigure}{t}{2.3in} % htbp
\centering
\vspace{-0.9cm}
\resizebox{2.3in}{!}{    
\includegraphics[width=\linewidth]{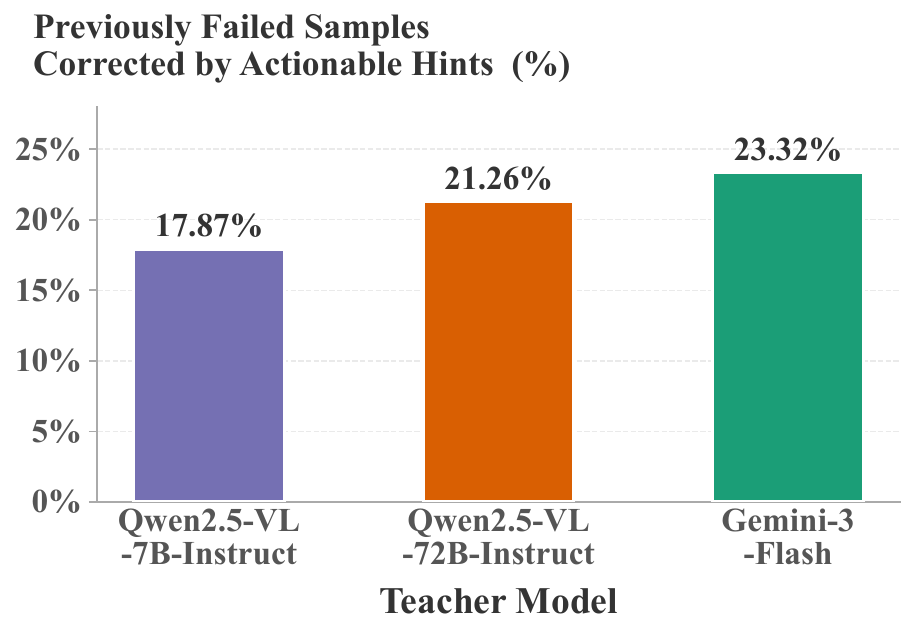}}
\caption{\textbf{Effectiveness of actionable hints from different teacher models.} 
Stronger teacher models produce more effective actionable hints, enabling a larger fraction of originally incorrect responses to be turned into correct ones.}
\vspace{-0.5cm}
\label{fig:second-rollout}
\end{wrapfigure}
analyze the role of teacher models in diagnostic analysis.
As shown in \cref{tab:example-part1,tab:example-part2}, different teachers produce noticeably different diagnoses for the same failure case.
In this example, the student model misinterprets the shaded region: Qwen2.5-VL-7B-Instruct gives a misleading diagnosis, Qwen2.5-VL-72B-Instruct correctly identifies the geometric relation, and Gemini-3-Flash provides the most fine-grained analysis.
This suggests that stronger teachers can provide more faithful diagnostic signals for Anchor Evolution.
\cref{fig:second-rollout} shows that stronger teachers generate more effective \texttt{actionable\_hint}s, which directly improves the success rate of guided second rollouts.
Since only trajectories that reach the correct ground truth are retained for training, a higher second-rollout success rate further improves the efficiency of verified data construction.
This trend is also reflected in \cref{tab:teacher-sft}: under the same data-construction time, the 7B self-teacher produces fewer verified samples and achieves an Avg. of 56.0, while the 72B teacher constructs more verified data and achieves an Avg. of 56.9.
When the 7B self-teacher is given more construction time to match the verified data size, its performance increases to 56.8 Avg., approaching the 72B teacher.
These results suggest that teacher strength affects AnE mainly through the efficiency and quality of verified data construction, rather than by directly transferring teacher reasoning trajectories.
Therefore, AnE is not strictly dependent on access to a stronger teacher, since the teacher only provides diagnoses, search keywords, and hints, while the final training trajectories are generated by the student and retained only after ground-truth verification.
Considering both performance and cost, Gemini-3-Flash achieves the best result but has substantially higher API cost, so we use Qwen2.5-VL-72B-Instruct as the default teacher in the main experiments.

\begin{table*}[t]
  \centering
  \small
\caption{\textbf{Comparison of teacher models in diagnostic analysis.}
All experiments use the same 40K seed RL dataset and 300K seed SFT dataset; we vary the teacher model and additionally report a matched-data-size setting for the 7B self-teacher.
Stronger teachers improve guided second-rollout success and verified data construction efficiency, which further affects downstream performance.
\textsuperscript{*} denotes API-based inference time, which is not directly comparable to local model deployment.}
  \label{tab:teacher-sft}
  \renewcommand{\arraystretch}{1.15}
  \resizebox{\textwidth}{!}{%
  \begin{tabular}{l|c|cccccccc|c}
    \toprule
    \textbf{Teacher Model} & \textbf{Data Size / Time} & \textbf{MathVista} & \textbf{MathVision} & \textbf{MathVerse} & \textbf{LogicVista} & \textbf{SciQA} & \textbf{MMMU} & \textbf{EMMA} & \textbf{MMSTAR} & \textbf{Avg} \\
\midrule
Preliminary RL 
& -- & 77.4 & 29.5 & 53.7 & 44.2 & 93.1 & 56.0 & 29.5 & 66.6 & 56.3 \\

\multirow{2}{*}{Qwen2.5-VL-7B-Instruct}
& 361K / 17.9h & 76.8 & 29.1 & \textbf{54.8} & 43.1 & 92.6 & 55.9 & 28.8 & 67.0 & 56.0 \\

& 429K / 28.8h & 76.3 & \textbf{32.1} & 52.1 & \textbf{48.0} & 92.2 & \textbf{57.0} & 30.2 & 66.2 & 56.8 \\

Qwen2.5-VL-72B-Instruct
& 429K / 17.9h & 76.4 & 32.0 & 53.2 & 45.5 & 93.4 & 55.0 & \textbf{30.7} & \textbf{68.5} & 56.9 \\
Gemini-3-Flash
& 451K / - \textsuperscript{*} & \textbf{78.0} & 32.0 & 53.9 & 46.7 & \textbf{93.5} & 56.1 & \textbf{30.7} & 67.4 & \textbf{57.3} \\
    \bottomrule
  \end{tabular}%
  }
  \vspace{0.5mm}
\end{table*}

\begin{table*}[t]
    \centering
    \caption{\textbf{Example of teacher-model comparison (Part I).} 
An example of an incorrect model response caused by misunderstanding the geometric relationship in the figure.}
    \scriptsize

    \begin{tcolorbox}[
        failureframe,
        title=\textcolor{white}{\textbf{Diagnostic Analysis by Different Teacher Models (Part I)}}
    ]

    \begin{minipage}[t]{0.18\linewidth}
        \vspace{0pt}
        \centering
        \includegraphics[width=0.58\linewidth]{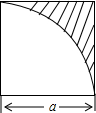}
    \end{minipage}
    \hfill
    \begin{minipage}[t]{0.80\linewidth}
        \vspace{0pt}
        \begin{tcolorbox}[infocard]
        \fontsize{6.5}{7.1}\selectfont
        \textbf{Question.} Calculate the shaded area, where $a=4$ cm.

        \vspace{2pt}
        \textbf{GT.} $16-4\pi$
        \end{tcolorbox}
    \end{minipage}

    \vspace{2pt}

    \begin{tcolorbox}[
        responsecard,
title=\textbf{Model's Incorrect Response},        fonttitle=\bfseries
    ]
    \fontsize{6.5}{7.1}\selectfont\ttfamily
<think>
To calculate the area of the shaded part, we need to consider the geometric shapes involved. The shaded area appears to be composed of a quarter of a circle and a right triangle.

1. Area of the Quarter Circle: the radius $r$ of the circle is given as $a=4$ cm; the area of a full circle is $\pi r^2$; therefore, the area of a quarter circle is $\frac{1}{4}\pi r^2$.

2. Area of the Right Triangle: the right triangle has one side equal to the radius of the circle, $r=4$ cm; the other side of the triangle is also $4$ cm (since it's a square cut in half). The area of a right triangle is given by $\frac{1}{2}\times \text{base}\times \text{height}$.

3. Combining the Areas: the shaded area is the sum of the area of the quarter circle and the area of the right triangle. Substitute $r=4$ cm into the formulas.

Let's calculate each part step by step.

Area of the quarter circle:
$\text{Area}=\frac{1}{4}\pi r^2=\frac{1}{4}\pi(4)^2=\frac{1}{4}\pi\times16=4\pi\ \text{cm}^2$

Area of the right triangle:
$\text{Area}=\frac{1}{2}\times4\times4=\frac{1}{2}\times16=8\ \text{cm}^2$

Total shaded area:
$\text{Shaded area}=\text{Area of the quarter circle}+\text{Area of the right triangle}=4\pi+8\ \text{cm}^2$

Now, we need to calculate the numerical value:
$4\pi\approx4\times3.14159=12.56636\ \text{cm}^2$

So,
$\text{Shaded area}\approx12.56636+8=20.56636\ \text{cm}^2$

Therefore, the final area of the shaded part is:
$\boxed{4\pi+8}$
</think>
$\boxed{4\pi+8}$
    \end{tcolorbox}

    \end{tcolorbox}

    \label{tab:example-part1}
\end{table*}
\begin{table*}[t]
    \centering
    \caption{\textbf{Example of teacher-model comparison (Part II).} 
Diagnostic analyzes from Qwen2.5-VL-7B-Instruct, Qwen2.5-VL-72B-Instruct, and Gemini-3-Flash for the same incorrect model response.}
    \scriptsize

    \begin{tcolorbox}[
        failureframe,
        title=\textcolor{white}{\textbf{Diagnostic Analysis by Different Teacher Models (Part II)}}
    ]

    \begin{tcolorbox}[
        innersubbox,
title=\textbf{Comparison of Diagnostic Analyzes},        fonttitle=\bfseries
    ]

    {\fontsize{6.1}{6.5}\selectfont
    \begin{tcolorbox}[
        analysiscard,
        title=\textbf{Qwen2.5-VL-7B-Instruct as Teacher},
        left=0.7mm,
        right=0.7mm,
        top=0.5mm,
        bottom=0.5mm
    ]
    \textbf{Error Analysis.} The model incorrectly combined the areas of the quarter circle and the right triangle, leading to an incorrect total shaded area. The model calculated the area of the quarter circle as \(4\pi\) ~cm$^2$ and the area of the right triangle as 8 ~cm$^2$, then added them together to get \(4\pi + 8\) ~cm$^2$. However, the correct approach should subtract the area of the right triangle from the area of the quarter circle, as the right triangle is not part of the shaded region but rather lies within the quarter circle.

    \textbf{Primary Error.}

    \textbf{Type.} \textit{Reasoning / Strategy}.

    \textbf{Explanation.} The model failed to correctly subtract the area of the right triangle from the area of the quarter circle, which is necessary to find the area of the shaded part.

    \textbf{Actionable Hint.} Strategy: Subtract the area of the right triangle from the area of the quarter circle. Mechanism: Calculate the area of the quarter circle as \(4\pi\) ~cm$^2$ and the area of the right triangle as 8 ~cm$^2$, then subtract 8 ~cm$^2$ from 4$\pi$ ~cm$^2$.

    \textbf{Search Keywords.} \texttt{"incorrectly added areas", "quarter circle", "right triangle", "subtraction error"}.

    \textbf{Secondary Errors.} None.
    \end{tcolorbox}

    \begin{tcolorbox}[
        analysiscard,
        title=\textbf{Qwen2.5-VL-72B-Instruct as Teacher},
        left=0.7mm,
        right=0.7mm,
        top=0.5mm,
        bottom=0.5mm
    ]
    \textbf{Error Analysis.} The model's logic deviated when it incorrectly calculated the area of the shaded part. The model claimed the shaded area is the sum of the area of a quarter circle and the area of a right triangle. However, the image/text actually shows that the shaded area is the area of the square minus the area of a quarter circle. The model calculated 4$\pi$ + 8, but the correct derivation implies 16 - 4$\pi$.

    \textbf{Primary Error.}

    \textbf{Type.} \textit{Reasoning / Strategy}.

    \textbf{Explanation.} The model failed to correctly interpret the geometric relationship between the shapes, leading to an incorrect combination of areas.

    \textbf{Actionable Hint.} Strategy: Identify the relationship between shapes in the figure. Mechanism: Recognize that the shaded area is the area of the square minus the area of the quarter circle, and calculate accordingly: Area of square - Area of quarter circle = a² - (1/4)$\pi$a².

    \textbf{Search Keywords.} \texttt{"area of shaded part", "quarter circle and square", "calculate shaded area", "geometric shapes relationship"}.

    \textbf{Secondary Errors.} None.
    \end{tcolorbox}

    \begin{tcolorbox}[
        analysiscard,
        title=\textbf{Gemini-3-Flash as Teacher},
        left=0.7mm,
        right=0.7mm,
        top=0.5mm,
        bottom=0.5mm
    ]
    \textbf{Error Analysis.} The model correctly identifies the side length $a=4$ and calculates the area of a quarter-circle with radius 4 ($4\pi$). However, it fails to correctly perceive the geometric composition of the shaded region. The model claims: 'The shaded area appears to be composed of a quarter of a circle and a right triangle' and subsequently adds these areas together. In reality, the image shows a square of side $a$ where a quarter-circle is unshaded; the shaded area is the difference between the square and the quarter-circle ($a^2 - \frac{1}{4}\pi a^2 = 16 - 4\pi$). The model failed to recognize the square as the bounding container and misidentified the shaded region as an additive composite of parts that do not exist in the figure.

    \textbf{Primary Error.}

    \textbf{Type.} \textit{Perception / Understanding}.

    \textbf{Explanation.} The model misidentified the spatial relationship between the shapes. It failed to perceive that the shaded region is a 'remainder' (Square - Sector) and instead hallucinated an additive structure ('quarter circle + triangle').

    \textbf{Actionable Hint.} Strategy: Identify the 'Bounding Box' and 'Subtractive Shapes' first. Mechanism: Explicitly name the outermost polygon (e.g., square) and any internal cut-outs (e.g., quarter-circle), then determine if the shaded area is the intersection, union, or difference of these primitives.

    \textbf{Search Keywords.} \texttt{"calculate the area of the shaded part", "area of shaded region square minus quarter circle", "shaded area of a square with a quarter circle removed", "geometry problem find area of shaded region", "area between square and inscribed sector"}.

    \textbf{Secondary Errors.} None.
    \end{tcolorbox}

    }

    \end{tcolorbox}
    \end{tcolorbox}

    \label{tab:example-part2}
\end{table*}

\paragraph{Generalization to Another Model Family.}
To examine whether AnE generalizes beyond Qwen2.5-VL, we further apply AnE to InternVL3.5-2B-Instruct~\cite{wang2025internvl3}.
As shown in \cref{tab:internvl-generalization}, AnE brings consistent improvements over the reported InternVL3.5-2B baseline and the Preliminary RL model.
The average score improves from 47.9 to 51.2 after Preliminary RL, and further increases to 57.2 after three AnE rounds.
This indicates that AnE can transfer to another model family and a smaller model scale.
\begin{table*}[t]
  \centering
  \small
  \caption{
  \textbf{Generalization to another model family.}
  We apply AnE to InternVL3.5-2B-Instruct and observe consistent gains across evolution rounds.
  }
  \label{tab:internvl-generalization}
  \renewcommand{\arraystretch}{1.15}
  \resizebox{\textwidth}{!}{%
  \begin{tabular}{l|cccccccc|c}
    \toprule
    \textbf{Method} & \textbf{MathVista} & \textbf{MathVision} & \textbf{MathVerse} & \textbf{LogicVista} & \textbf{SciQA} & \textbf{MMMU} & \textbf{EMMA} & \textbf{MMSTAR} & \textbf{Avg} \\
    \midrule
    InternVL3.5-2B-Instruct & 60.8 & 27.0 & 39.6 & 41.2 & \textbf{93.3} & 53.0 & 22.6 & 46.2 & 47.9 \\
    Preliminary RL & 67.9 & 33.0 & 38.7 & 45.5 & 90.1 & 54.6 & 28.4 & 51.3 & 51.2 \\
    AnE-$1^{st}$ & 71.3 & 40.0 & 42.1 & 47.1 & 92.0 & \textbf{57.9} & 29.8 & 55.6 & 54.5 \\
    AnE-$2^{nd}$ & 72.9 & 41.0 & 45.5 & \textbf{48.9} & 92.4 & 56.0 & 32.9 & 61.5 & 56.4 \\
    AnE-$3^{rd}$ & \textbf{74.8} & \textbf{42.5} & \textbf{48.1} & 46.9 & 92.2 & 56.8 & \textbf{34.5} & \textbf{61.9} & \textbf{57.2} \\
    \bottomrule
  \end{tabular}%
  }
\end{table*}

\paragraph{Additional Evaluation on Broader Reasoning Tasks.}
To further evaluate the generalization of AnE on more benchmarks, we conduct additional experiments on MME~\cite{fu2026mme}, OmniSpatial~\cite{jia2025omnispatial}, and UI-Vision~\cite{nayak2025ui}, covering commonsense, spatial, and GUI-oriented multimodal reasoning, respectively.
As shown in \cref{tab:additional-eval}, AnE achieves clear gains on MME and OmniSpatial, demonstrating its effectiveness on commonsense and spatial reasoning tasks.
The improvement on UI-Vision is relatively limited, likely due to the limited GUI-related data in the seed training set.
\begin{table}[t]
  \centering
  \small
  \caption{
  \textbf{Additional evaluation on broader multimodal reasoning tasks.}
  AnE improves commonsense and spatial reasoning performance, while the gain on GUI reasoning is relatively limited.
  }
  \label{tab:additional-eval}
  \renewcommand{\arraystretch}{1.15}
  \resizebox{0.55\linewidth}{!}{%
  \begin{tabular}{lccc}
    \toprule
    \textbf{Method} & \textbf{MME$_{\text{sum}}$} & \textbf{OmniSpatial} & \textbf{UI-Vision} \\
    \midrule
    Qwen2.5-VL-7B-Instruct & 2347.0 & 39.2 & 0.9 \\
    AnE-$1^{st}$ & 2418.8 & 43.4 & 2.3 \\
    AnE-$2^{nd}$ & 2392.3 & 43.8 & 1.1 \\
    AnE-$3^{rd}$ & \textbf{2457.5} & \textbf{45.5} & \textbf{4.0} \\
    \bottomrule
  \end{tabular}%
  }
\end{table}

\paragraph{Evaluation Variance Analysis.}
To assess evaluation stability, we repeat the testing of key variants three times.
The results show that the 8-benchmark average has small variance and the ranking among methods remains stable.
As shown in \cref{tab:evaluation-variance}, AnE-$1^{st}$ consistently outperforms Preliminary RL, the variant with only Truth Anchor Expansion, and the variant with only Scaffold-Stripping.
The standard deviation of the average score is at most 0.18\%, suggesting that the observed gains are unlikely to be caused by evaluation randomness.

\begin{table*}[t]
  \centering
  \scriptsize
\caption{
  \textbf{Three random repeated evaluations.}
  We additionally repeat the evaluation of key ablation variants three times and report accuracy (\%).
  The Mean row reports the mean $\pm$ standard deviation across the three runs.
}
  \label{tab:evaluation-variance}
  \renewcommand{\arraystretch}{1.12}
  \resizebox{\textwidth}{!}{%
  \begin{tabular}{ll|cccccccc|c}
    \toprule
    \textbf{Setting} & \textbf{Run} 
    & \textbf{MathVista} & \textbf{MathVision} & \textbf{MathVerse} & \textbf{LogicVista} 
    & \textbf{SciQA} & \textbf{MMMU} & \textbf{EMMA} & \textbf{MMSTAR} & \textbf{Avg.} \\
    \midrule
    \multirow{4}{*}{Preliminary RL}
    & 1 & 77.30 & 32.57 & 53.35 & 45.09 & 92.53 & 57.78 & 29.84 & 66.87 & 56.91 \\
    & 2 & 78.00 & 32.93 & 53.53 & 45.76 & 92.36 & 57.33 & 30.06 & 67.27 & 57.15 \\
    & 3 & 78.20 & 32.73 & 53.71 & 45.54 & 92.71 & 57.78 & 30.20 & 67.13 & 57.25 \\
    & Mean 
    & 77.83$\pm$0.47 & 32.74$\pm$0.18 & 53.53$\pm$0.18 & 45.46$\pm$0.34 
    & 92.53$\pm$0.18 & 57.63$\pm$0.26 & 30.03$\pm$0.18 & 67.09$\pm$0.20 
    & 57.10$\pm$0.17 \\
    \midrule

    \multirow{4}{*}{w/ Truth Anchor}
    & 1 & 77.60 & 36.28 & 56.32 & 48.44 & 93.49 & 57.78 & 30.09 & 68.00 & 58.50 \\
    & 2 & 78.70 & 36.55 & 56.17 & 48.88 & 93.52 & 56.67 & 30.34 & 68.20 & 58.63 \\
    & 3 & 78.60 & 36.41 & 56.45 & 49.55 & 93.63 & 57.56 & 30.20 & 67.53 & 58.74 \\
    & Mean 
    & 78.30$\pm$0.61 & 36.41$\pm$0.14 & 56.31$\pm$0.14 & 48.96$\pm$0.56 
    & 93.55$\pm$0.07 & 57.34$\pm$0.59 & 30.21$\pm$0.13 & 67.91$\pm$0.34 
    & 58.62$\pm$0.12 \\
    \midrule

    \multirow{4}{*}{w/ Scaffold-Stripping}
    & 1 & 77.90 & 35.39 & 58.71 & 44.20 & 93.00 & 56.56 & 30.77 & 68.47 & 58.12 \\
    & 2 & 77.50 & 34.90 & 58.78 & 45.31 & 93.11 & 55.56 & 30.56 & 68.53 & 58.03 \\
    & 3 & 77.90 & 35.33 & 58.40 & 43.97 & 92.88 & 56.33 & 30.70 & 68.00 & 57.94 \\
    & Mean 
    & 77.77$\pm$0.23 & 35.21$\pm$0.27 & 58.63$\pm$0.20 & 44.49$\pm$0.72 
    & 93.00$\pm$0.12 & 56.15$\pm$0.52 & 30.68$\pm$0.11 & 68.33$\pm$0.29 
    & 58.03$\pm$0.09 \\
    \midrule

    \multirow{4}{*}{AnE-$1^{st}$}
    & 1 & 80.50 & 40.10 & 60.61 & 47.99 & 93.47 & 58.00 & 30.42 & 68.00 & 59.89 \\
    & 2 & 79.70 & 39.90 & 60.61 & 50.00 & 93.82 & 57.78 & 30.77 & 68.53 & 60.14 \\
    & 3 & 79.90 & 39.77 & 60.66 & 49.11 & 93.73 & 56.78 & 30.85 & 68.40 & 59.90 \\
    & Mean 
    & 80.03$\pm$0.42 & 39.92$\pm$0.17 & 60.63$\pm$0.03 & 49.03$\pm$1.01 
    & 93.67$\pm$0.18 & 57.52$\pm$0.65 & 30.68$\pm$0.23 & 68.31$\pm$0.28 
    & 59.98$\pm$0.14 \\
    \bottomrule
  \end{tabular}%
  }
\end{table*}
\paragraph{Comparison between Text-only and Vision-Language Retrieval.}
In the main experiments, we use Qwen3-Embedding-8B as the default retriever for Truth Anchor Expansion.
Since this retriever is text-only, it does not directly encode visual content in the multimodal pool.
To analyze this design choice, we further conduct an ablation using Qwen3-VL-Embedding-8B~\cite{li2026qwen3vlembed} as a vision-language retriever.
As shown in \cref{tab:vl-retriever}, the VL retriever slightly improves AnE-$1^{st}$ from 59.9\% to 60.1\% on average.
However, it increases the database-side embedding time from 8h to 21h.
Considering the marginal performance gain and the substantially higher preprocessing cost, we adopt the text-only retriever as the default setting.

\begin{table}[t]
  \centering
  \small
  \caption{
    \textbf{Ablation on retrieval encoder.}
    Vision-language retrieval yields only marginal improvement but requires much higher database-side embedding cost.
  }
  \label{tab:vl-retriever}
  \renewcommand{\arraystretch}{1.15}
  \resizebox{0.55\linewidth}{!}{%
  \begin{tabular}{lcc}
    \toprule
    \textbf{Retriever} & \textbf{Avg. of AnE-$1^{st}$} & \textbf{Embedding Time} \\
    \midrule
    Qwen3-Embedding-8B & 59.9 & 8h \\
    Qwen3-VL-Embedding-8B & 60.1 & 21h \\
    \bottomrule
  \end{tabular}%
  }
\end{table}

\paragraph{Sensitivity Analysis on Rollout Number \(K\).} The rollout number \(K\) controls
\begin{wrapfigure}{r}{0.52\linewidth}
    \centering
    \vspace{-2mm}
    \includegraphics[width=\linewidth]{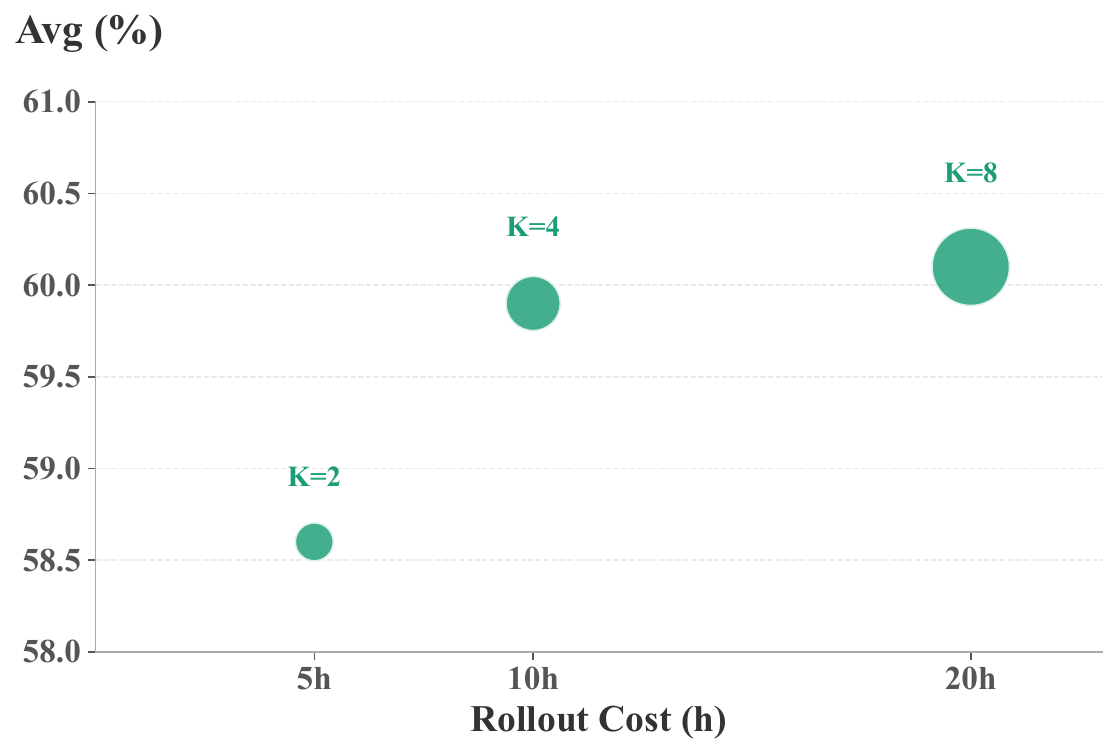}
        \caption{
    \textbf{Sensitivity analysis on rollout number \(K\).}
    \(K=2\) reduces rollout cost but leads to lower performance, while \(K=8\) brings only marginal improvement over \(K=4\) with doubled rollout cost.
    }
    \label{fig:k-ablation}
    \vspace{-3mm}
\end{wrapfigure}
 the reliability of data routing in Failing-Frontier Discovery.
A smaller \(K\) reduces computation but may lead to noisy partitioning of redundant, volatile, and frontier samples, while a larger \(K\) provides more stable routing at a higher rollout cost.
As shown in \cref{fig:k-ablation}, using \(K=2\) reduces the rollout cost to 5.0 hours but decreases the average performance to 58.6\%.
Increasing \(K\) from 4 to 8 only brings a marginal improvement from 59.9\% to 60.1\%, while doubling the rollout cost from 10.0 to 20.0 hours.
Therefore, we set \(K=4\) as the default choice, which provides the best trade-off between routing reliability and computational cost.

\end{document}